\documentclass[sigconf]{acmart}
\AtBeginDocument{%
  }

\setcopyright{acmlicensed}
\copyrightyear{2026}
\acmYear{2026}
\acmDOI{10.1145/pending}
\acmConference[KDD '26]{32nd ACM SIGKDD Conference on Knowledge Discovery and Data Mining}{August 9--13, 2026}{Jeju, Korea}
\acmISBN{978-1-4503-XXXX-X/26/08}
\usepackage{graphicx}
\usepackage{svg}
\usepackage[T1]{fontenc}
\usepackage{listings}

\DeclareUnicodeCharacter{0394}{\ensuremath{\Delta}}

\newif\ifsupp
\suppfalse

\begin{document}

\title{BenchBench: Benchmarking Automated Benchmark Generation}

\author{Yandan Zheng}
\affiliation{%
  \institution{Nanyang Technological University}
  \city{Singapore}
  \country{Singapore}}
\email{yandan.zheng@ntu.edu.sg}

\author{Haoran Luo}
\authornote{Co-corresponding authors.}
\affiliation{%
  \institution{Nanyang Technological University}
  \city{Singapore}
  \country{Singapore}}
\email{haoran.luo@ieee.org}

\author{Zhenghong Lin}
\affiliation{%
  \institution{Nanyang Technological University}
  \city{Singapore}
  \country{Singapore}}
\email{hongzhenglin970323@gmail.com}

\author{Wenjin Liu}
\affiliation{%
  \institution{Nanyang Technological University}
  \city{Singapore}
  \country{Singapore}}
\email{wenjinliu23@outlook.com}

\author{Luu Anh Tuan}
\authornotemark[1]
\affiliation{%
  \institution{Nanyang Technological University}
  \city{Singapore}
  \country{Singapore}}
\email{anhtuan.luu@ntu.edu.sg}

\renewcommand{\shortauthors}{Zheng et al.}

% \begin{abstract}
% Benchmarks remain the primary means of tracking progress in large language models (LLMs), yet they increasingly struggle to remain fresh, trustworthy, and diagnostic. Creating new benchmarks is costly, static test sets are vulnerable to contamination, and scalable scoring for open-ended items often relies on LLM judges, which can introduce systematic biases and prompt sensitivity. 

% We argue that evaluation should therefore extend beyond “how well models answer benchmarks” to “how well models design them.” We propose to evaluate LLMs not only as answerers, but as benchmark designers: a three-stage pipeline extracts domain cards from seed datasets, prompts multiple models to generate controlled benchmark suites, and validates them with a panel of answerers to build a designer–answerer matrix.

% This setup makes benchmark generation measurable: good designers produce valid, constraint-faithful items that discriminate among models and expose systematic interactions such as format effects, modality/language fidelity, and self/family bias—providing a scalable lens on the meta-capability of benchmark design. 

% \end{abstract}

\begin{abstract}
Benchmarks are the de facto standard for tracking progress in large language models (LLMs), yet static test sets can rapidly saturate, become vulnerable to contamination, and are costly to refresh. Scalable evaluation of open-ended items often relies on LLM judges, introducing additional sources of bias and prompt sensitivity. We argue that evaluation must extend beyond \emph{how well models answer benchmarks} to \emph{how well models design them}. We introduce \textbf{BenchBench}, a three-stage pipeline and dataset for benchmarking automated benchmark generation: (i) extract structured \emph{domain cards} from seed benchmarks, (ii) prompt multiple designer LLMs to generate quota-controlled suites, and (iii) validate items with a multi-model answerer panel using exact/numeric/symbolic verifiers when possible and rubric-guided judging otherwise, yielding designer--answerer matrices with item-level quality flags and psychometric diagnostics. Across nine variants spanning computer science, mathematics, medicine, and theory-of-mind reasoning (including multilingual and multimodal settings), we generate 16.7K items, retain \(\approx\)15K core items post-filtering, and produce \(\approx\)152K graded model--item responses. BenchBench shows that benchmark-design ability is only moderately correlated with answer-time strength (Spearman \(\rho\approx0.37\)), invalidity is negatively associated with discrimination (Pearson \(r\approx-0.62\)), and the resulting designer--answerer matrices enable scalable audits of format/modality/language fidelity and suite-dependent self/family interactions. The project is available at: https://github.com/koanatakiyo/BenchBench.
\end{abstract}




\maketitle

\begin{figure*}
  \centering
  \includegraphics[width=0.88\linewidth]{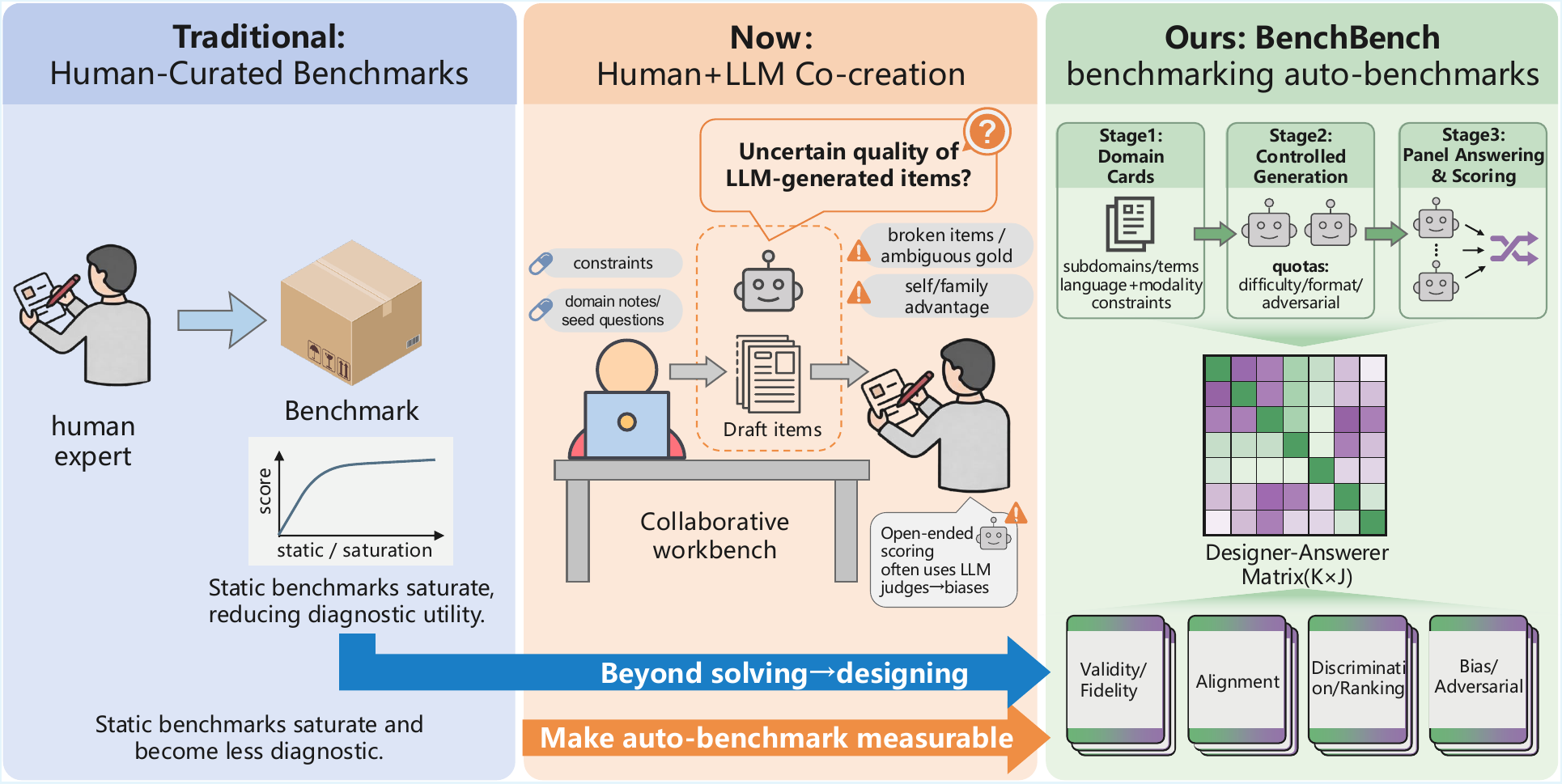}
  \caption{Benchmark creation is shifting from expert curation to human+LLM co-creation; BenchBench benchmarks the next step: \emph{auto-benchmarks}. It makes generated suites measurable via domain cards, quota-controlled generation, and panel-based validation, yielding designer--answerer matrices for auditing validity, diagnostic utility, and interaction effects. Saturation trends adapted from \cite{hendrycks2021mmlu, epoch2025brief}.}
  \label{fig:teaser}
\end{figure*}

\section{Introduction}

Large language models (LLMs) \cite{team2024gemini, openai2025gpt5, openai2024gpt4, deepseek2025v3, meta2025llama4, bytedance2025doubao, qwen2025qwen3, claudesonnet45, xai2025grok41} are advancing at a pace that routinely outstrips our ability to evaluate them. Benchmarks and leaderboards remain the de facto measure of progress, yet they face three structural pressures: \textbf{(i) saturation}, \textbf{(ii) cost}, and \textbf{(iii) self-referential evaluation}. Static test sets are rapidly optimized against, shrinking headroom even as live benchmarks attempt to refresh content \cite{zeng2025futurex, white2025livebench, hendrycks2020measuring}. High-quality benchmark creation requires domain expertise and iterative refinement to eliminate ambiguity and shortcuts \cite{liang2022holistic, srivastava2023beyond, li2025active}. Finally, pipelines increasingly rely on LLMs to generate and judge items; judging is prompt-sensitive and biased, and contamination further blurs generalization versus memorization \cite{white2025livebench, wang2023self, zeng2024automatic, geng2024large, panickssery2024llm, deng-etal-2024-investigating}. These trends make systematic meta-evaluation of \emph{benchmark quality} urgent \cite{perlitz2024these, qian2026benchmark}.

These trends motivate a paradigm shift: beyond measuring how well models \emph{answer} benchmark questions, we should measure how well models \emph{design} benchmarks. Benchmark design is not merely "question writing"—it requires selecting appropriate skills to evaluate, controlling difficulty and format, anticipating failure modes, and producing items that discriminate among models without introducing artifacts that advantage particular model families or evaluation heuristics. Crucially, design ability is a meta-capability: it assesses whether a model understands not only what is correct but also what is \emph{informative to test}. Currently, benchmark proliferation proceeds without rigorous meta-evaluation \cite{perlitz2024these, qian2026benchmark}, making systematic assessment of generation quality increasingly urgent.

In this work, we propose a three-stage, multi-domain pipeline that treats LLMs as both designers and answerers. (i) \textbf{Domain card extraction}: From four seed benchmarks spanning computer science (CSBench \cite{song2024cs}), mathematics (WeMath \cite{qiao2025we}), medicine (MedXpertQA \cite{zuo2025medxpertqa}), and theory-of-mind reasoning (ToMBench \cite{chen2024tombench}), we derive structured domain cards that summarize subdomains, key terminology, and modality/language constraints. (ii) \textbf{Controlled generation}: Multiple designer LLMs generate new benchmark suites under explicit quotas for difficulty tiers, adversarial intent proportion, and question format diversity. (iii) \textbf{Panel-based validation}: A diverse panel of answerer models attempts all generated items, with scoring prioritized as follows: objective verifiers (exact, numeric, and symbolic matching) whenever possible, and rubric-guided LLM judging only when objective verification is unavailable. We then analyze the resulting designer–answerer performance matrix using psychometric-style metrics (difficulty and discrimination), ranking consistency, and behavioral signals (format usage, visual/language fidelity, and self/family bias).

Our framing makes benchmark generation measurable: a designer is effective if the resulting items remain valid and well-specified, preserve intended constraints (domain, language, modality), and retain diagnostic headroom and discrimination. Because generation conditions on domain cards rather than raw seed items, the same pipeline can also instantiate \emph{private-domain} evaluations from proprietary knowledge representations without exposing underlying corpora. Our main contributions are:
\begin{itemize}
\item \textbf{Benchmarking benchmark makers.} We formulate LLM benchmark design as a measurable task and evaluate designers via a designer$\times$answerer response matrix with item-level quality flags.
\item \textbf{BenchBench: a multi-domain instantiation and artifact.} We instantiate the task across computer science, mathematics, medicine, and theory-of-mind reasoning, including multilingual and multimodal variants, and release the resulting artifacts for reproducible meta-evaluation.
\item \textbf{Objective-first validation and analysis.} We prioritize exact/numeric/symbolic verifiers and invoke rubric-guided judging only when necessary, enabling psychometric diagnostics (difficulty/discrimination), ranking preservation checks, and audits of format/modality/language fidelity and self/family interactions.
\end{itemize}

% We address the following research questions (RQ1--RQ8):

% \begin{enumerate}
%     \item Which LLMs design better benchmarks, and does model strength correlate with design ability?
%     \item Do designers show format/style preferences, and how do these affect item discrimination?
%     \item How do designers handle visual references in multimodal domains?
%     \item Do generated suites preserve established model rankings?
%     \item Do we observe self- or family-level biases in the designer–answerer matrix?
%     \item Can designers create effective adversarial and “model-stumping” items?
%     \item Do designers respect language constraints in multilingual settings?
%     \item How do design capabilities vary across domains?
% \end{enumerate}

We address eight research questions: 
\textbf{(RQ1)} which LLMs design better benchmarks and whether model strength correlates with design ability; \textbf{(RQ2)} format/style preferences and their impact on discrimination; \textbf{(RQ3)} handling of visual references in multimodal domains; \textbf{(RQ4)} preservation of established model rankings; \textbf{(RQ5)} self- and family-level biases; \textbf{(RQ6)} creation of effective adversarial and model-stumping items; \textbf{(RQ7)} respect for language constraints in multilingual settings; and \textbf{(RQ8)} variation of design capabilities across domains.

\section{Task Definition}
\label{sec:task}

We formalize \emph{automated benchmark design} as a measurable task. For each dataset variant $D$ (domain $\times$ language $\times$ modality), we extract a structured \emph{domain card} $C_D$ from a seed benchmark, prompt a \emph{designer} LLM to synthesize a quota-controlled suite, and validate the suite with a fixed \emph{answerer} panel. The resulting designer--answerer response matrix enables systematic evaluation of benchmark-design quality, including validity/fidelity, specification alignment, diagnostic utility, and interaction effects.

\paragraph{\textbf{Notation.}}
Let $\mathcal{V}$ be the set of dataset variants. For each $D\in\mathcal{V}$, let $\mathcal{I}^{\text{seed}}_{D}$ denote the seed (human-curated) items and $C_D$ the extracted domain card. Let $\mathcal{G}$ be a set of designer models and $\mathcal{A}$ a set of answerer models (these sets may overlap in practice).

For each $(D,g)$ with $g\in\mathcal{G}$, Stage~2 generates a suite
$\mathcal{I}^{\text{gen}}_{D,g}=\{i\}_{i=1}^{N_D}$, where each item
$i=(x_i,y_i,m_i)$ contains the question $x_i$, a reference answer $y_i$, and metadata $m_i$ (e.g., question type, declared difficulty, standard/adversarial tag, and language/modality tags).
Stage~3 filters $\mathcal{I}^{\text{gen}}_{D,g}$ into a \emph{core} set $\mathcal{I}^{\text{core}}_{D,g}\subseteq \mathcal{I}^{\text{gen}}_{D,g}$ and collects responses $r_{a,i}$ for each $a\in\mathcal{A}$ and $i\in\mathcal{I}^{\text{core}}_{D,g}$.

\paragraph{\textbf{Objective-first scoring.}}
Each response is scored using an objective-first hierarchy. When an objective verifier is available, we obtain hard correctness $c_{a,i}\in\{0,1,\bot\}$; otherwise, a rubric-guided soft score $s_{a,i}\in[0,1]\cup\{\bot\}$. We define the final per-response score
\begin{equation}
z_{a,i}=
\begin{cases}
c_{a,i}, & c_{a,i}\neq \bot,\\
s_{a,i}, & c_{a,i}= \bot \ \wedge\ s_{a,i}\neq \bot,\\
\bot, & \text{otherwise.}
\end{cases}
\end{equation}
Let $\mathcal{I}^{\text{core}}_{D,g,a}=\{i\in\mathcal{I}^{\text{core}}_{D,g}\mid z_{a,i}\neq\bot\}$. The designer--answerer matrix for variant $D$ is
\begin{equation}
M^{D}_{a,g}=\frac{1}{|\mathcal{I}^{\text{core}}_{D,g,a}|}\sum_{i\in\mathcal{I}^{\text{core}}_{D,g,a}} z_{a,i}.
\end{equation}
This matrix, together with item metadata and quality flags, is the substrate for all downstream analyses.

\section{Method}
\label{sec:method}

% BenchBench operationalizes benchmark design as a closed-loop pipeline with three stages: (Stage~1) domain-card construction, (Stage~2) quota-controlled generation by designer LLMs, and (Stage~3) panel-based validation that produces designer--answerer matrices and item-level quality signals.
\begin{figure*}[t]
  \centering
  \includegraphics[width=\textwidth]{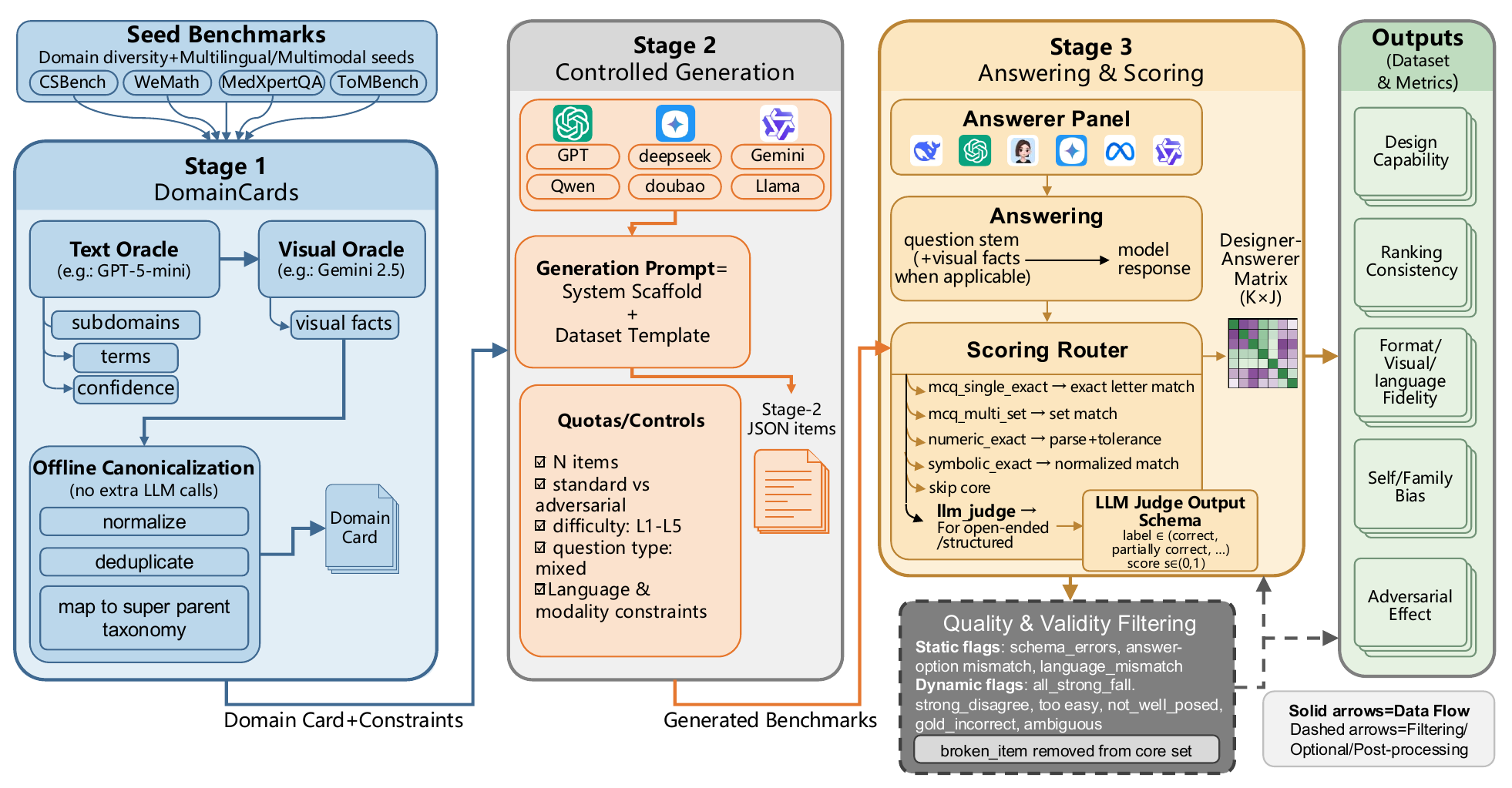}
  \caption{\textbf{BenchBench knowledge-guided pipeline.}
Stage~1 extracts deterministic domain cards from seed benchmarks using text (and vision when applicable) oracles plus offline canonicalization.
Stage~2 generates quota-controlled benchmark suites conditioned on the domain card and closes coverage gaps via deficit-driven top-ups.
Stage~3 performs post-hoc cleaning, runs a multi-model answerer panel, routes scoring via exact/numeric/symbolic matching or LLM judging, and applies static/dynamic quality gates to produce a core designer$\times$answerer response matrix and downstream evaluation metrics.}
  \label{fig:pipeline}
\end{figure*}

BenchBench operationalizes benchmark design as an end-to-end, \emph{measurable} process (Fig.~\ref{fig:pipeline}). Starting from diverse seed benchmarks, we extract a compact domain representation ($C_D$) that anchors generation to subdomains, terminology, and language/modality constraints. Designers then synthesize quota-controlled suites under explicit coverage and style controls. Finally, a fixed answerer panel validates each suite under an objective-first scoring hierarchy, producing a dense designer--answerer response matrix with item-level quality flags. This matrix view is crucial: it lets us separate \emph{invalid/ill-posed} items from \emph{hard but diagnostic} items, and it exposes systematic interactions (format effects, modality/language fidelity, and self/family biases) that are invisible in single-score evaluations.

\subsection{Stage 1: Domain Card Construction}
\label{sec:method_stage1}

For each variant $D$, Stage~1 extracts a structured \emph{domain card} $C_D$ from the seed items $\mathcal{I}^{\text{seed}}_{D}$. The domain card serves as a controlled knowledge representation for generation rather than directly reusing seed questions. Concretely, $C_D$ summarizes (i) an ontology of coarse topics/subdomains (``super-parents''), (ii) representative terms and constraints, and (iii) modality/language requirements. For multimodal seeds, the card includes explicit visual-grounding guidance (e.g., what constitutes valid visual reference) to support controlled comparisons between text-only and visual-primed generation.
We release all domain cards as YAML files (with rendered hierarchy plots) under \texttt{domain\_cards/} in the artifact; examples are provided in the supplementary material.

\subsection{Stage 2: Quota-Controlled Generation (LLMs as Designers)}
\label{sec:method_stage2}

Given $C_D$, each designer $g\in\mathcal{G}$ generates a suite $\mathcal{I}^{\text{gen}}_{D,g}$ under explicit quotas over (i) super-parent coverage, (ii) question type/format, (iii) declared difficulty tiers, and (iv) standard vs.\ adversarial intent. Generation is performed in batches; after each batch, we compute quota deficits and perform targeted ``top-up'' batches until the suite reaches the desired coverage. Each item includes a required schema (question fields + metadata) so that downstream validation is deterministic and auditable.

\paragraph{\textbf{Item axes}}
Each generated item stores three control variables in metadata: \textbf{(i) Design intent:} \texttt{design\_type} $\in$ \{\texttt{standard}, \texttt{adversarial}\}   \textbf{(ii) Reasoning depth:}  \texttt{declared\_difficulty} $\in$ \{\texttt{L1}, \dots, \texttt{L5}\} \textbf{(iii) Response format:} \texttt{question\_type} $\in$ \{\texttt{mcq\_single}, \texttt{mcq\_multi}, \dots, \texttt{judgment}\}. \texttt{adversarial} denotes \emph{intent} (failure-mode targeting) and is orthogonal to depth: adversarial items can be shallow-but-tricky, while standard items can be deep multi-step reasoning. 
The \texttt{L1}--\texttt{L5} tiers capture reasoning depth (from recall/single-step to multi-step, multi-concept). 
Format affects both the interaction surface and the scoring regime: MCQ variants are hard-scored by exact/set match, while open/structured items may require rubric-guided judging (Section~\ref{sec:method_stage3}).

\subsection{Stage 3: Panel Validation, Scoring, and Core Filtering}
\label{sec:method_stage3}

Stage~3 validates each generated suite using a fixed answerer panel $\mathcal{A}$ and produces (i) response traces and scores, (ii) item-level quality flags, and (iii) aggregated designer--answerer matrices (Section~\ref{sec:task}). We prioritize objective verifiers whenever applicable; otherwise we invoke rubric-guided LLM judging.

\paragraph{\textbf{Scoring hierarchy.}}
We score responses in the following order:
(1) exact-match for MCQ,
(2) numeric/symbolic matching for short answers when applicable,
(3) rubric-guided LLM judging for remaining open-ended/structured responses,
(4) \texttt{skip\_core} for unverifiable or ill-posed cases.
In parallel, a dynamic quality pass uses panel signals and a dedicated quality judge to flag items that are ambiguous, have incorrect gold answers, or violate constraints. Items failing static schema checks or dynamically flagged are excluded from the core set.

\subsection{Evaluation Protocol: Four Metric Pillars}
\label{sec:pillars}

We organize metrics into four complementary pillars.

\paragraph{\textbf{P1: Validity \& fidelity.}}
Quantify validity via the non-core rate:
\begin{equation}
R_{\text{noncore}}(D,g)=1-\frac{|\mathcal{I}^{\text{core}}_{D,g}|}{|\mathcal{I}^{\text{gen}}_{D,g}|}.
\end{equation}
For judged subsets, we also report judge-flag rates (e.g., \texttt{ambiguous}, \texttt{gold\_incorrect}), capturing semantic/factual fidelity beyond schema compliance.

\paragraph{\textbf{P2: Specification alignment.}}
We measure deviations from Stage~2 targets derived from $C_D$ (domain coverage, format mix, difficulty mix, and language/modality constraints). We compute divergences between realized and target distributions (e.g., L1/JS divergence for domain/format), correlations between declared difficulty and empirical difficulty, and compliance rates for language and visual-use constraints (text-only vs.\ visual-primed variants).

\paragraph{\textbf{P3: Diagnostic utility and ranking reliability.}}
On hard-scored core items, we compute item difficulty
$p(i)=\mathbb{E}_{a\in\mathcal{A}}[c_{a,i}]$
and discrimination via corrected item--total correlation (point-biserial), summarized as designer-level means.
To assess whether generated suites preserve established answerer rankings, we compare rankings induced by each designer suite against a reference ranking (seed- or matrix-derived), using Kendall’s $\tau$ with bootstrap confidence intervals over items.

\paragraph{\textbf{P4: Interaction effects and adversarial behavior.}}
To audit systematic designer--answerer interactions, we report self and family gaps. Let $F(\cdot)$ map a model to a provider family. Family advantage for answerer $a$ is:
\begin{equation}
\Delta_{\text{fam}}(D,a)=
\mathbb{E}_{g\in\mathcal{G}_{F(a)}}[M^{D}_{a,g}]
-
\mathbb{E}_{g\notin\mathcal{G}_{F(a)}}[M^{D}_{a,g}].
\end{equation}

We also evaluate adversarial effectiveness as the drop in strong-panel accuracy between standard and adversarial subsets, while requiring that retained adversarial items remain core, which makes them valid and scorable.

\section{Experimental Setup}
\label{sec:exp}

\subsection{BenchBench overview}
\label{sec:exp_overview}

We instantiate the task in Section~\ref{sec:task} on nine dataset variants spanning computer science, mathematics, medicine, and theory-of-mind reasoning, including multilingual (EN/ZH) and multimodal settings. For multimodal seeds, we evaluate two controlled generation conditions (\texttt{textonly} and \texttt{visualprimed}) to isolate the effect of visual grounding in generated items. Across all variants, we generate 16,669 unique items and retain 14,893 core items after filtering, producing 170,580 graded responses in total (152,275 on the core set). About 22\% of (core) responses require rubric-guided judging due to open-ended or structured formats; the remainder are scored by objective verifiers.
We summarize the pooled realized composition (Items vs.\ Core) over intent (\texttt{standard} vs.\ \texttt{adversarial}), declared difficulty (L1--L5), and question types in the supplementary artifact appendix ~\ref{app:artifact}.

\begin{table}[t]
\centering
\small
\caption{\textbf{BenchBench overview (evaluated variants).}
\#Items/\#Core count unique generated questions aggregated across designers within each variant. CSBench-EN uses an extended designer/answerer panel; all other variants use 6 designers and 10 answerers. Shorthand dataset labels correspond to full identifiers as follows: MedXpertQA (Text) is medxpertqa\_text; MedXpertQA (MM-T) is medxpertqa\_mm\_stage2\_textonly; MedXpertQA (MM-V) is medxpertqa\_mm\_stage2\_visualprimed; WeMath (Text) is wemath\_stage2\_textonly; and WeMath (Visual) is wemath\_stage2\_visualprimed.}
\label{tab:benchbench_overview}
\begin{tabular}{lccc}
\toprule
\textbf{Variant} & \textbf{\#Designers} & \textbf{\#Answerers} & \textbf{\#Core / \#Items} \\
\midrule
CSBench (EN) & 8 & 12 & 2,051 / 2,390 \\
CSBench (ZH) & 6 & 10 & 1,643 / 1,780 \\
ToMBench (EN) & 6 & 10 & 1,594 / 1,790 \\
ToMBench (ZH) & 6 & 10 & 1,513 / 1,796 \\
MedXpertQA (Text) & 6 & 10 & 1,684 / 1,802 \\
MedXpertQA (MM-T)) & 6 & 10 & 1,645 / 1,800 \\
MedXpertQA (MM-V) & 6 & 10 & 1,555 / 1,761 \\
WeMath (Text) & 6 & 10 & 1,658 / 1,788 \\
WeMath (Visual) & 6 & 10 & 1,550 / 1,762 \\
\bottomrule
\end{tabular}

\end{table}

\subsection{Model Panels}
\label{sec:exp_panels}

\paragraph{\textbf{Designers.}}
We use a common designer set of six models across all variants:
\textit{gpt-5-mini}, \textit{gemini-2.5-flash}, \textit{deepseek-chat},
\textit{qwen3-next-80b-a3b-instruct}, \textit{doubao-seed-1.6-flash}, and \textit{llama-4-maverick}.
For CSBench-EN only, we additionally include \textit{claude-sonnet-4.5} and \textit{grok-4.1-fast} as designers to broaden family coverage.

\paragraph{\textbf{Answerers and judges.}}
The answerer panel includes ten models:
\textit{gpt-5-mini}, \textit{gpt-4.1-mini}, \textit{gemini-2.5-flash}, \textit{gemini-2.0-flash},
\textit{deepseek-v3.2-chat}, \textit{qwen3-next-80b-a3b-instruct}, \textit{qwen3-vl-flash},
\textit{doubao-seed-1.6-flash}, \textit{llama-4-maverick}, and \textit{llama-3.3-70b-instruct}.
CSBench-EN additionally includes \textit{claude-sonnet-4.5} and \textit{grok-4.1-fast} as answerers.
The rubric-guided judge for open-ended scoring is \textit{gemini-2.5-pro}; a separate quality judge is invoked for ambiguity/gold checks on a subset of items.

\subsection{Generation and Validation Hyperparameters}
\label{sec:exp_hparams}

Stage~2 generation uses temperature $T=0.8$ by default (unless overridden per provider), with fixed schemas and batch-wise top-ups to satisfy quotas. Stage~3 answering is deterministic ($T=0$) for all answerers. Judge decoding is deterministic ($T=0$) with a fixed rubric prompt. We enforce provider-specific concurrency limits and batch sizes; the complete configuration is included in the released artifact.
For transparency, we also provide the Stage~1--3 prompt templates (system scaffolds and per-dataset templates) in the supplementary material.

\subsection{Baselines and Reference Rankings}
\label{sec:exp_baselines}

We use three reference baselines in analysis:
(1) quota targets derived from domain cards (alignment),
(2) seed- or matrix-derived consensus rankings for ranking-preservation tests, and
(3) leave-one-out cross-designer aggregation to contextualize self/family interaction effects.

\paragraph{\textbf{Human validity audit.}}
To complement automatic schema checks, dynamic flags, and judge-based adjudication, we conduct a two-rater human audit on a stratified sample of \(\approx\)150 generated items (sampled across designers and quality strata). Raters score ambiguity, key correctness, solvability, and format compliance (Likert 1--5) and mark a binary \texttt{fatal\_flaw} indicator; agreement and correlations with automatic metrics are reported Appendix~\ref{tab:human_audit_vs_auto}.

\section{Results}
\label{sec:results}

\paragraph{\textbf{Scope and measurement.}}
We report results for the nine evaluated variants summarized in Section~\ref{sec:exp_overview}. Stage~3 applies static schema/constraint checks and dynamic quality filtering to form the \emph{core} set (Section~\ref{sec:method_stage3}); items removed from core are reflected by Broken\% (non-core rate). Unless noted otherwise, item difficulty $p(i)$ and discrimination are computed on core items with hard correctness from objective verifiers; rubric-judged items remain in matrix aggregates but are excluded from hard-score psychometrics.

\begin{table}[t]
\centering
\caption{Aggregated designer leaderboard across all variants (core set). Broken\% is the non-core rate (lower is better); MeanDiscr is mean item discrimination (point-biserial) on hard-scored core items (higher the better). Full metrics are reported in the supplementary material Table \ref{tab:designer_leaderboard}.}
\begin{tabular}{lrr}
\toprule
Designer & Broken\% & MeanDiscr \\
\midrule
gpt-5-mini & \textbf{3.5} & \textbf{0.301} \\
llama-4-maverick & 10.9 & 0.241 \\
deepseek-chat & 7.8 & 0.213 \\
doubao-seed-1-6-flash-250828 & 12.6 & 0.209 \\
gemini-2.5-flash & 5.0 & 0.205 \\
qwen3-next-80b-a3b-instruct & 14.1 & 0.189 \\
\bottomrule
\end{tabular}

\label{tab:designer_leaderboard}
\end{table}

\begin{figure}[t]
    \centering
    \includegraphics[width=\linewidth]{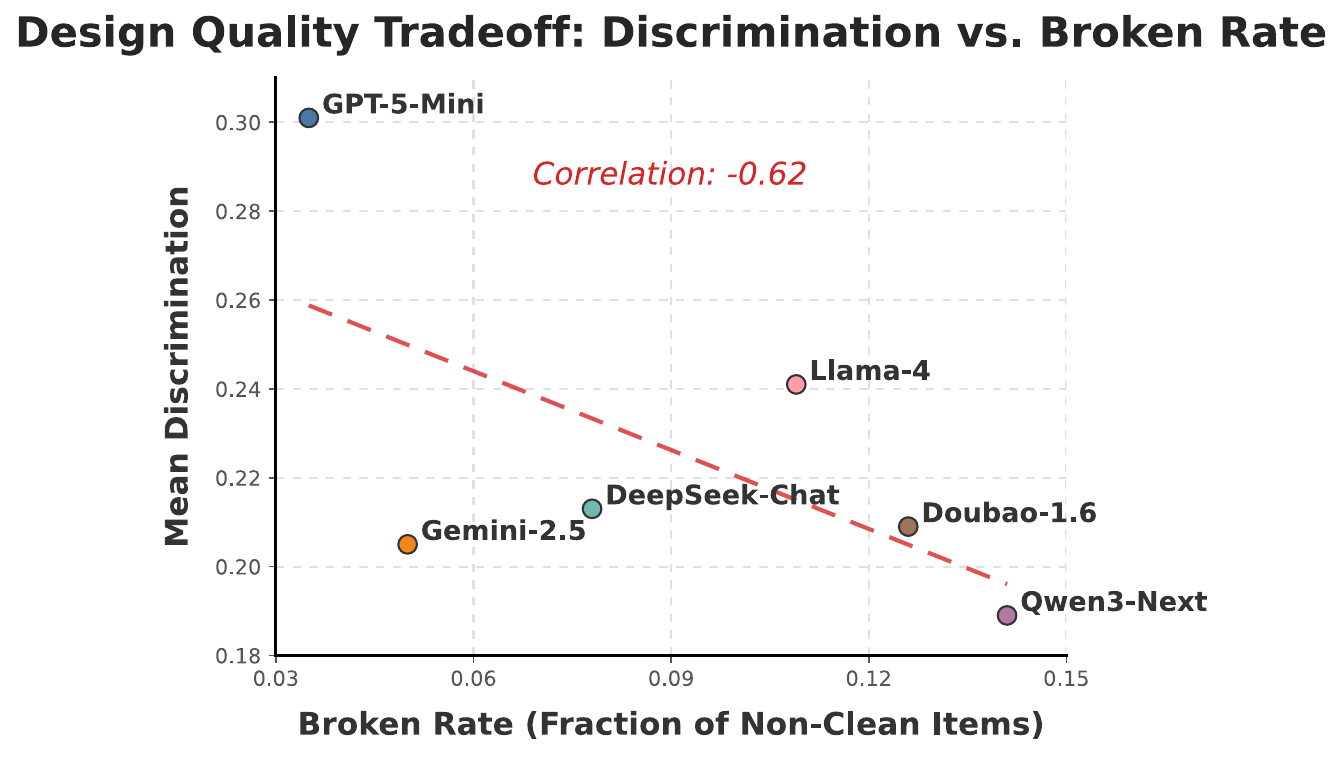}
    \caption{Validity--discrimination tradeoff across designers (pooled across variants). Broken\% is the non-core rate; MeanDiscr is averaged over hard-scored core items.}
    \label{fig:tradeoff}
\end{figure}

\subsection{RQ1: Which LLMs design better benchmarks? Does model strength correlate with design ability?}
Designer quality varies along multiple axes. In the pooled leaderboard, designers differ substantially in Broken\%, mean discrimination, and the share of negative-discrimination items (Table~\ref{tab:designer_leaderboard}). Figure~\ref{fig:tradeoff} shows a clear coupling between validity and diagnosticity: designers with higher Broken\% tend to yield lower mean discrimination (Pearson $r\approx-0.62$), suggesting that many suites that appear ``hard'' do so through invalidity or ambiguity rather than calibrated separation among answerers. We also observe cross-variant reversals in designer ordering, so aggregate leaderboards do not fully determine per-domain performance.

Design ability is only moderately associated with a model’s answering strength. A strength--design scatter on models appearing in both roles shows a positive but imperfect relationship (Spearman $\rho\approx0.37$), consistent with benchmark making behaving as a distinct meta-capability rather than a direct proxy of task accuracy.

\subsection{RQ2: Do designers exhibit format/style preferences, and how do these affect discrimination?}
Designers realize different format mixes under quota controls (e.g., MCQ vs.\ open/structured). These differences co-occur with changes in scoring coverage and filtering: suites that lean more heavily on open/structured formats increase reliance on rubric-guided scoring and expand the surface for ambiguity, while MCQ suites concentrate mass on hard-verifiable items. As a result, discrimination and difficulty measured on the retained hard-scored core set reflect both item quality and realized format composition (Table~\ref{tab:designer_leaderboard}).

\subsection{RQ3: How do designers handle visual references in multimodal domains?}
We compare \texttt{textonly} v.s. \texttt{visualprimed} generation in multimodal domains. For MedXpertQA-MM, visual priming shifts both mean discrimination and mean difficulty relative to text-only generation (Fig.~\ref{fig:visual_delta_med}), indicating that enabling visual grounding can change the diagnostic regime of generated suites. In contrast, WeMath exhibits smaller average shifts under priming (reported in the supplementary material). Across variants, these shifts can be accompanied by changes in core retention, reflecting additional failure modes such as underspecified or inconsistent visual instructions (Table~\ref{tab:benchbench_overview}).

\begin{figure}[t]
    \centering
    \includegraphics[width=\linewidth]{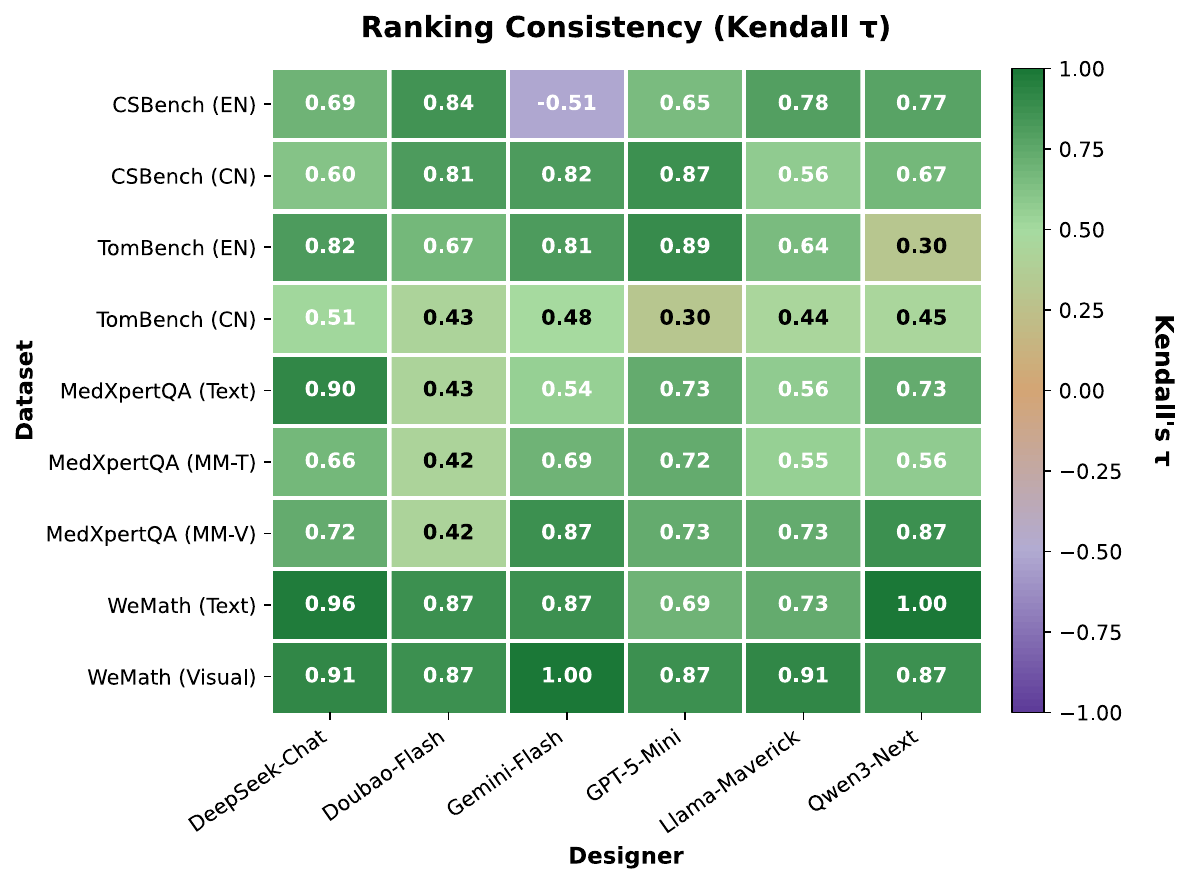}
    \caption{Ranking preservation across variants (Kendall's $\tau$), comparing designer-induced answerer rankings against a reference ranking.  Higher values indicate greater preservation of established model order.Variant labels follow Table~\ref{tab:benchbench_overview}.}
    \label{fig:ranking_tau}
\end{figure}

\subsection{RQ4: Do generated suites preserve established model rankings?}
We measure ranking preservation via Kendall’s $\tau$ between the answerer ranking induced by each designer suite and a reference/consensus ranking. Figure~\ref{fig:ranking_tau} shows that ranking consistency varies sharply by variant: some settings yield stable rankings across designers, whereas others are more sensitive to ambiguity and reduced discrimination. This variation aligns with the variant-dependent diagnostic regime visible in discrimination and core retention.

\begin{figure*}[htbp]
    \centering
    \includegraphics[width=\textwidth]{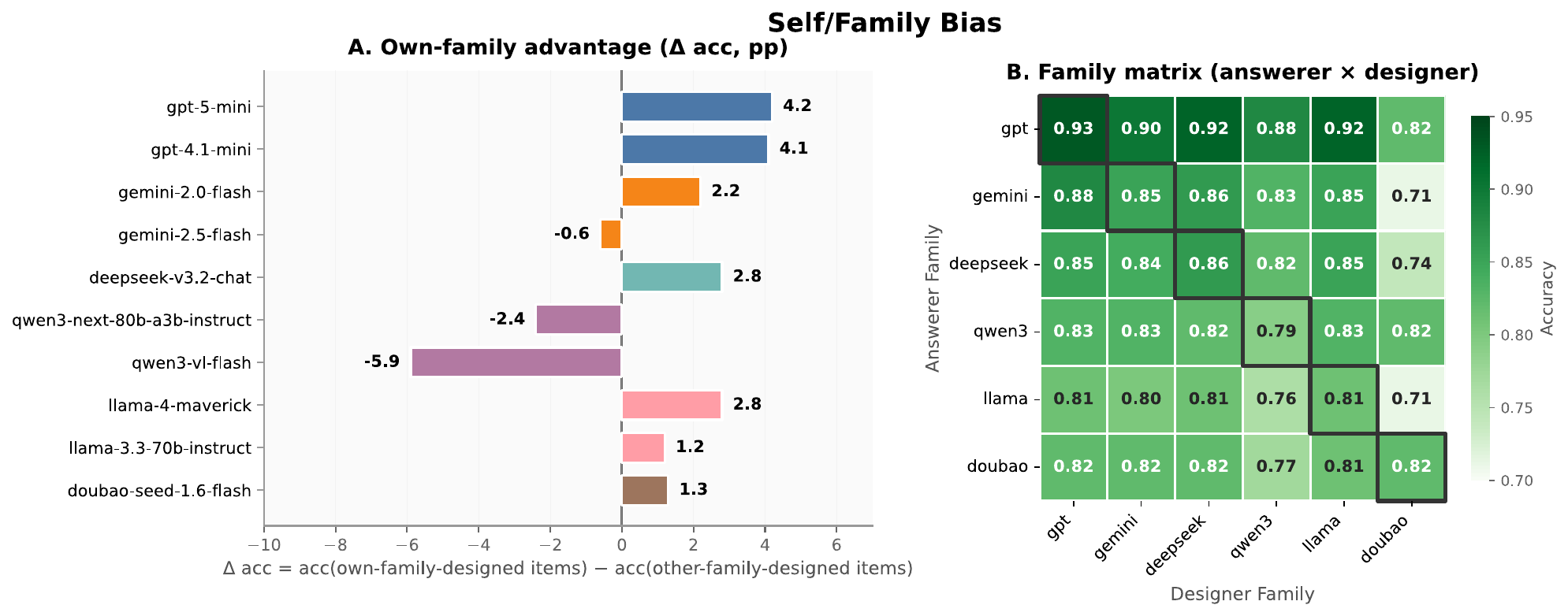}
    \caption{Self/family bias in the designer--answerer matrix. Left: own-family advantage (accuracy on own-family items minus accuracy on other-family items, percentage points). Right: family-level accuracy matrix (darker = higher accuracy). Cells are aggregated over answerers; rows/columns grouped by model family.}
    \label{fig:family_bias}
\end{figure*}

\subsection{RQ5: Do we observe self- or family-level biases in the designer--answerer matrix?}
The designer--answerer matrix enables direct auditing of same-family interactions. Figure~\ref{fig:family_bias} summarizes own-family versus other-family gaps and the family-level matrix aggregated over suites. Family effects are measurable but typically modest in aggregate, with variant- and family-specific deviations. Because CSBench-EN is the only variant with additional families (Claude and Grok) in both designer and answerer panels, we report an extended-family CSBench-EN case study in the supplementary material to avoid changing the family universe in pooled comparisons.

\begin{figure}[t]
    \centering
    \includegraphics[width=1\linewidth]{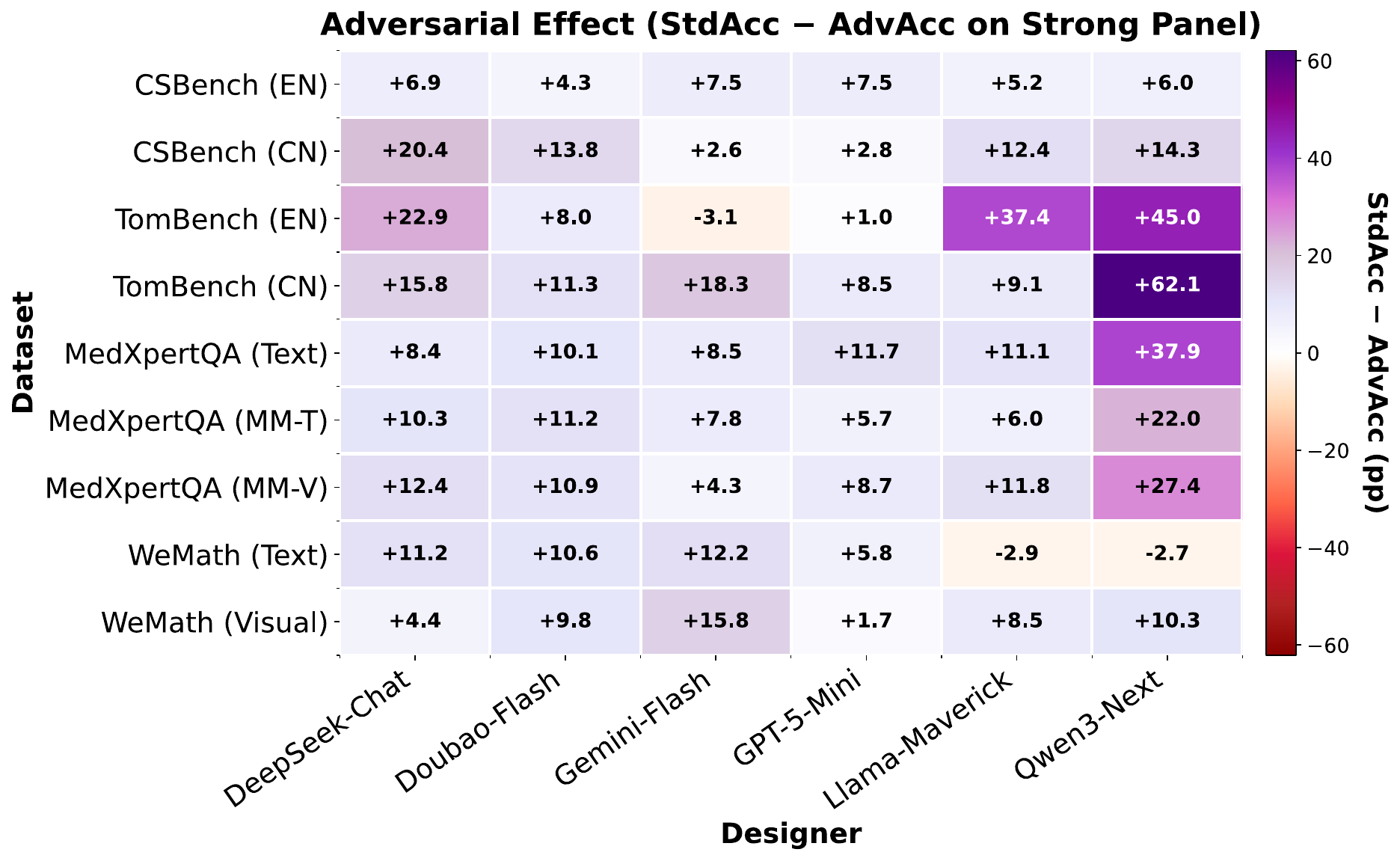}
    \caption{Adversarial effect on strong-panel accuracy. Each cell shows the drop in mean accuracy (standard minus adversarial subset, the higher the better adversarial design) on the strong answerer subset. e.g., $+$62.1\% for Qwen3-Next on TomBench (CN) indicates effective adversarial design.}
\label{fig:adv_effect}

\end{figure}

\subsection{RQ6: Can designers create effective adversarial and model-stumping questions?}
Adversarial subsets are intended to reduce strong-panel accuracy while remaining valid and diagnostic. Figure~\ref{fig:adv_effect} reports the drop in strong-panel accuracy from standard to adversarial items, showing substantial designer$\times$variant heterogeneity. These drops are most informative when adversarial items remain core and retain non-trivial discrimination, rather than being filtered as ill-posed.

\subsection{RQ7: Do designers respect language constraints in multilingual settings?}
For bilingual variants (CSBench and ToMBench), generated suites largely follow the target language at the content level, but metadata conventions and edge cases require normalization for consistent scoring and reporting. We therefore treat language fidelity as a combination of content compliance and schema/metadata compliance, and report detailed compliance summaries and common violations in the supplementary material and released artifact.

\subsection{RQ8: How do design capabilities vary across domains?}
Design capability is not uniform across domains and variants. We observe substantial cross-variant shifts in core retention, difficulty headroom, discrimination, and ranking stability (Table~\ref{tab:benchbench_overview}, Fig.~\ref{fig:ranking_tau}). These shifts co-occur with cross-variant changes in designer ordering, indicating that benchmark-making performance depends on domain structure and modality/language constraints rather than being a single transferable scalar.

% Across variants, BenchBench exposes benchmark-making as measurable along complementary axes—validity, format realization, diagnostic utility (difficulty/discrimination), ranking stability, and systematic interaction effects—revealing substantial designer$\times$domain heterogeneity.

Across different variants, BenchBench demonstrates that benchmark creation can be quantitatively assessed on multiple complementary dimensions—including validity, format compliance, diagnostic utility (such as difficulty and discrimination), ranking consistency, and systematic interaction effects—highlighting significant variability between designers and domains.

\section{Analysis \& Discussion}
\label{sec:analysis}

BenchBench is designed to benchmark \emph{benchmark makers}: the goal is not to maximize any single score, but to characterize how different designers trade off validity, controllability, and diagnostic utility under a fixed validation instrument (the answerer panel and scoring router). This section synthesizes the main mechanisms behind the results in Section~\ref{sec:results} and highlights practical implications for building and auditing auto-generated benchmarks.

\paragraph{\textbf{Validity Is a Precondition for Diagnosticity}}
A consistent pattern is that suites with higher non-core (Broken\%) rates tend to exhibit lower discrimination on the retained hard-scored core set (Fig.~\ref{fig:tradeoff}). This coupling suggests that many ``hard-looking'' suites are hard for the wrong reason: ambiguity, inconsistent constraints, or unstable gold answers introduce noise that weakens item--ability relationships and collapses discrimination. Practically, this makes filtering outcomes first-class signals of designer competence rather than a mere preprocessing detail. Reporting Broken\% alongside discrimination (Table~\ref{tab:designer_leaderboard}) helps distinguish calibrated difficulty from difficulty-by-corruption.

\paragraph{\textbf{Benchmark-Making Is Related to, but Not Determined by, Benchmark-Solving}}
Designer performance is only moderately associated with answerer strength. This separation is expected: writing diagnostic, well-posed items requires specification-following, anticipating failure modes across heterogeneous answerers, and producing questions that are informative \emph{as measurements}, not merely solvable. The moderate association supports treating benchmark design as a distinct meta-capability that should be evaluated directly rather than inferred from task accuracy.

\paragraph{\textbf{Format and Scoring Coverage Are Coupled Measurement Levers}}
Format is not just presentation; it changes the measurement instrument. Even under quota controls, designers realize different format mixes, which affects (i) the share of items that can be objectively verified versus rubric-judged, and (ii) the surface for ambiguity and dynamic filtering. Consequently, discrimination measured on the hard-scored core set reflects both item quality and realized format composition (Table~\ref{tab:designer_leaderboard}). This motivates reporting format mix and the judged-vs-hard scoring split alongside psychometric metrics, especially when comparing designers that differ in open/structured usage.

\paragraph{\textbf{Multimodal Priming Changes the Diagnostic Regime}}
In multimodal medicine, visual priming shifts both difficulty and discrimination relative to text-only generation (Fig.~\ref{fig:visual_delta_med}), indicating that enabling visual grounding can reduce saturation and change what the suite measures. However, the same mechanism can introduce additional validity risks (e.g., underspecified visual instructions), which is reflected in variant-level core retention (Table~\ref{tab:benchbench_overview}). We therefore interpret multimodal gains jointly with validity signals: priming is most useful when it increases diagnostic utility without inflating non-core rates.

\paragraph{\textbf{Ranking Stability Serves as a Sanity Check for Benchmark Reliability}}
Benchmarks are often consumed as rankings. Figure~\ref{fig:ranking_tau} shows that ranking preservation varies sharply by variant, consistent with the variant-dependent validity/discrimination regime. In practice, ranking instability is a warning sign: it often co-occurs with reduced discrimination or increased filtering, suggesting the suite may be sensitive to ambiguity or saturation. Treating Kendall's $\tau$ as a first-class metric helps separate ``hard but diagnostic'' suites from ``hard but noisy'' ones.

\paragraph{\textbf{Matrix-Based Evaluation Exposes Interaction Effects Beyond Leaderboards}}
The designer--answerer matrix enables analyses that single-score benchmarks cannot: it reveals systematic interactions such as self/family effects (Fig.~\ref{fig:family_bias}). In aggregate these effects are typically modest, but variant- and family-specific deviations exist and can confound designer comparisons if not audited. CSBench-EN provides an extended-family case study (Claude/Grok) illustrating that own-family effects are not universally positive and that adding families changes the comparison universe; we therefore report that analysis separately in the supplementary material and keep pooled figures on the shared family intersection.

\paragraph{\textbf{Adversarial Effectiveness Requires Validity and Discrimination}}
Adversarial subsets can reliably reduce strong-panel accuracy (Fig.~\ref{fig:adv_effect}), but adversarial ``hardness'' is only informative when adversarial items remain core and retain non-trivial discrimination. This distinguishes adversarial design skill (creating precise, difficult, discriminative items) from adversarial failure modes (creating ambiguous or ill-posed questions that are filtered). Joint reporting of adversarial drops with Broken\% and discrimination is therefore necessary to avoid rewarding noise.

\paragraph{\textbf{Implications for Benchmarking Benchmark Makers}}
Our results suggest several practical guidelines for evaluating auto-benchmark generators:
\begin{itemize}
\item \textbf{Report multi-objective quality.} Combine validity (non-core rate), diagnostic utility (discrimination and headroom), and reliability (ranking stability) rather than a single leaderboard score.
\item \textbf{Make the instrument visible.} Report format mix and the fraction of objective-verifier versus judged scoring, since they change what is measured and how reproducible it is.
\item \textbf{Prefer matrix views to single numbers.} The designer--answerer matrix supports bias audits and interaction analyses that can prevent over-interpreting family- or judge-dependent artifacts.
\item \textbf{Interpret hardness conditionally.} Treat ``hard'' as progress only when it co-occurs with low non-core rates and non-trivial discrimination.
\end{itemize}

Overall, BenchBench shows that benchmarking benchmark makers is feasible at scale, but only when evaluation is treated as a controlled measurement pipeline that separates validity from difficulty, and summarizes outcomes at the level of suites and matrices rather than individual items.

\section{Related work}

\textbf{Benchmark Challenges and Meta-Evaluation.}
Progress in LLMs has exposed critical flaws in benchmarks, including saturation of static datasets and widespread data contamination \cite{ravaut2025a, golchin2024time}.
Dynamic approaches like LiveBench \cite{white2025livebench} mitigate contamination with temporally fresh, objectively scored questions and regular updates.
Meta-evaluation frameworks, notably LLM benchmarks agreement \cite{perlitz2024these} and Benchmark$^2$ \cite{qian2026benchmark}, begin to systematically assess benchmark quality itself.
Benchmark$^2$ introduces three complementary metrics—cross-benchmark ranking consistency, discriminability scores, and capability-alignment deviation (flagging intra-family upsets)—and demonstrates their use to identify unreliable suites and enable efficient selective subsetting across 15 existing benchmarks.
These efforts, however, remain retrospective: they diagnose quality in \emph{existing} (primarily human-crafted) benchmarks and do not assess LLMs' ability to proactively design new ones.

\medskip

\textbf{LLM-Generated Benchmarks.}
To address cost and contamination, recent works automate benchmark creation with LLMs.
Examples include AutoCodeBench \cite{chou2025autocodebench} for multilingual code synthesis via sandboxes, OSS-Bench \cite{jiang2025oss} from open-source repositories, and FrontendBench \cite{zhu2025frontendbench} for domain-specific execution-based tasks.
While effective for scalable, fresh item generation, these pipelines typically lack closed-loop validation of the resulting benchmarks' discriminative power, artifact resistance, modality/language fidelity, or behavioral biases across diverse models.

While prior meta-evaluation frameworks \cite{perlitz2024these, qian2026benchmark} provide valuable post-hoc diagnostics for existing datasets, \textbf{BenchBench} extends meta-evaluation to \emph{LLM-generated} benchmarks. We apply related ideas (e.g., ranking consistency and intra-family analysis) to designer-induced performance matrices but add generation controls, multimodal/language fidelity checks, and behavioral metrics for self/family bias. This treats benchmark design as a measurable meta-capability, enabling direct ranking of LLMs as designers—insights unattainable from purely retrospective or generative approaches.

\section{Limitations \& Future Work}
\label{sec:limitations}

While BenchBench provides a systematic framework for evaluating LLMs as benchmark designers, several limitations constrain the scope and generalizability of our findings.

% Finally, several limitations remain and define future directions.

\textbf{Seed dependence:} domain cards inherit the coverage and biases of seed benchmarks; broader and more diverse seeds should improve robustness.
% \textbf{Novelty and contamination:} while we deduplicate and filter, we cannot fully rule out training memorization or near-duplicates of public items without stronger retrieval-based checks.

\textbf{Judge and open-ended reliability:} expanding objective verifiers (executors, solvers) and incorporating multi-judge/human calibration would improve confidence.

\textbf{Generalization to new domains:} extending domain cards beyond the current seeds (and quantifying how much seed data is required) is an important next step for turning BenchBench into a general-purpose benchmark-construction toolkit.

% \textbf{}

\textbf{Number of panels:} The answerer panel consists of only 10 models, which may not fully capture performance variance across a broader range of systems, particularly smaller or specialized models. 

% \textbf{Dynamic quality:} Dynamic quality signals and strong-subset analyses rely on this fixed panel, potentially introducing sampling bias in difficulty estimates and discrimination metrics. 

\textbf{Domain coverage:} Domain coverage is restricted to four primary areas (computer science, mathematics, medicine, theory-of-mind), with seven variants for modality and language control. This leaves many important domains unexamined, such as physics, law, or creative writing, and limits conclusions about cross-domain transfer. 

% The adversarial and model-stomping items are identified via panel behavior but lack large-scale human validation to confirm validity or true "stumping" nature.

% Future work should address these constraints through:
% \begin{enumerate}
%     \item  Expansion to larger, more diverse answerer panels, including open-source fine-tunes and non-frontier models.
%     \item Incorporation of stronger or ensemble judges (e.g., majority voting across multiple high-capability systems) to reduce judging noise.
%     \item Extension to additional domains, languages (beyond English/Chinese), and modalities (e.g., audio, video).
%     \item Integration with agentic workflows, where designers interact iteratively or use external tools during generation.
%     \item Exploration of continuous benchmark refreshment mechanisms, leveraging BenchBench to maintain evolving evaluation suites over time.
% \end{enumerate}

% %%
% %% The next two lines define the bibliography style to be used, and
% %% the bibliography file.
\bibliographystyle{ACM-Reference-Format}
\bibliography{sample-base,benchbench}

% \ifsupp
\appendix

\section{BenchBench Artifact Details}
\label{app:artifact}

This appendix documents the released BenchBench artifact. BenchBench is a \emph{benchmark for benchmarking benchmark makers}: its primary goal is to provide a reproducible evaluation substrate (domain cards, quota-controlled suites, validation traces, and designer--answerer matrices), rather than a training dataset.

\subsection{Components}
The artifact contains four layers:

\paragraph{\textbf{(A1) Domain cards.}}
For each variant $D$, we release a structured domain card $C_D$ (YAML) summarizing subdomains (super-parents), representative terms, and language/modality constraints extracted from seed benchmarks. Domain cards are deterministic to improve reproducibility.

\paragraph{(\textbf{A2) Stage~2 generated suites.}}
For each $(D,g)$, we release the generated suite $\mathcal{I}^{\text{gen}}_{D,g}$ as JSONL with strict schema and metadata fields (format, declared difficulty, standard/adversarial tag, and language/modality tags), together with quota reports and any top-up records.

\paragraph{\textbf{(A3) Stage~3 validation traces and quality flags.}}
For each variant, we release answerer responses, parsed answers, the scoring method, and item-level quality flags used to construct the core set. This includes both objective-verifier outcomes and rubric-guided judge outputs when required.

\paragraph{\textbf{(A4) Aggregated matrices and analysis-ready tables.}}
We release designer--answerer matrices $M^D$ (hard/soft as applicable), per-variant metrics, and analysis tables/figures used in the paper.

\subsection{Evaluated Variants and Scale}
We evaluate nine variants (Section~\ref{sec:exp_overview}). Across all variants, the artifact contains 16,669 unique generated items and retains 14,893 core items post-filtering, producing 170,580 graded responses in total (152,275 on the core set). About 22\% of core responses are rubric-judged due to open-ended or structured formats.

\begin{table*}[t]
\centering
\small
\caption{\textbf{Pooled suite composition (Items vs.\ Core).}
Items count unique generated questions across the nine evaluated variants (Table~\ref{tab:benchbench_overview}).
Core retains items with \texttt{item\_status}\(\in\)\{\texttt{clean}, \texttt{suspect\_static}\} and \texttt{scoring\_method}\(\neq\)\texttt{skip\_core}.
Percentages are within Items (\(N=16{,}669\)) and within Core (\(N=14{,}893\)).}
\begin{tabular}{lrrrr}
\toprule
\textbf{Value} & \multicolumn{2}{c}{\textbf{Items}} & \multicolumn{2}{c}{\textbf{Core}} \\
 & \textbf{n} & \textbf{\%} & \textbf{n} & \textbf{\%} \\
\midrule
\multicolumn{5}{l}{\textbf{Intent} (\texttt{design\_type})} \\
\texttt{standard} & 12{,}620 & 75.7 & 11{,}597 & 77.9 \\
\texttt{adversarial} & 4{,}038 & 24.2 & 3{,}289 & 22.1 \\
\texttt{other/invalid} & 11 & $<$0.1 & 7 & $<$0.1 \\
\addlinespace
\multicolumn{5}{l}{\textbf{Difficulty} (\texttt{declared\_difficulty})} \\
L1 & 2{,}272 & 13.6 & 2{,}174 & 14.6 \\
L2 & 4{,}072 & 24.4 & 3{,}819 & 25.6 \\
L3 & 5{,}350 & 32.1 & 4{,}784 & 32.1 \\
L4 & 3{,}421 & 20.5 & 2{,}872 & 19.3 \\
L5 & 1{,}554 & 9.3 & 1{,}244 & 8.4 \\
\addlinespace
\multicolumn{5}{l}{\textbf{Question type} (\texttt{question\_type})} \\
\texttt{mcq\_single} & 10{,}324 & 61.9 & 9{,}708 & 65.2 \\
\texttt{open\_ended} & 3{,}366 & 20.2 & 2{,}869 & 19.3 \\
\texttt{mcq\_multi} & 1{,}397 & 8.4 & 1{,}121 & 7.5 \\
\texttt{structured} & 1{,}439 & 8.6 & 1{,}094 & 7.3 \\
\texttt{judgment} & 141 & 0.8 & 99 & 0.7 \\
\texttt{other/invalid} & 2 & $<$0.1 & 2 & $<$0.1 \\
\bottomrule
\end{tabular}

\label{tab:pooled_composition}
\end{table*}

\subsection{Human Validity Audit}
Two independent raters audited a stratified sample of 150 items (25 per designer across the six shared designers), spanning multiple variants and quality strata. Each item was annotated for ambiguity, key correctness, solvability, and format compliance (Likert 1--5) and a binary \texttt{fatal\_flaw} label. On the 146 items with complete two-rater labels, observed agreement on \texttt{fatal\_flaw} was 97.3\% and Krippendorff's \(\alpha=0.32\) (nominal). Agreement on the four Likert dimensions was low (Krippendorff's \(\alpha\) with ordinal distance; Table~\ref{tab:human_audit_agreement}), consistent with rater calibration differences and ceiling effects; we therefore treat Likert ratings as qualitative diagnostics and focus on \texttt{fatal\_flaw} for quantitative validation. The union fatal-flaw rate was 3.4\% (5/146). Across designers, \texttt{fatal\_flaw} rates show positive but inconclusive correlations with automatic brokenness and negative-discrimination rates due to small sample size (Spearman \(\rho=0.46\) vs.\ broken\%, bootstrap 95\% CI [\(-0.57\), 1.00]; \(\rho=0.40\) vs.\ neg-discr\%, CI [\(-0.60\), 1.00]).

\begin{table}[t]
\centering
\small
\caption{\textbf{Inter-rater agreement in E4 (Krippendorff's \(\alpha\), \(N=146\)).}
We compute \(\alpha\) for \texttt{fatal\_flaw} using nominal distance, and for the Likert (1--5) dimensions using ordinal distance. Negative \(\alpha\) indicates systematic disagreement and can arise under scale compression or rater calibration differences.}
\label{tab:human_audit_agreement}
\begin{tabular}{lcc}
\toprule
\textbf{Label} & \textbf{Scale} & \textbf{\(\alpha\)} \\
\midrule
\texttt{fatal\_flaw} & nominal (0/1) & 0.32 \\
ambiguity & ordinal (1--5) & -0.88 \\
key correctness & ordinal (1--5) & 0.02 \\
solvability & ordinal (1--5) & -0.43 \\
format compliance & ordinal (1--5) & -0.22 \\
\bottomrule
\end{tabular}

\end{table}

\begin{table}[t]
\centering
\small
\caption{\textbf{Human audit vs.\ automatic metrics (per designer).}
\texttt{fatal\_flaw} uses the union-of-raters definition; automatic metrics are computed over all generated suites (per-designer summaries are included in the released artifact).}
\label{tab:human_audit_vs_auto}
\begin{tabular}{lrrrr}
\toprule
\textbf{Designer} & \textbf{n} & \textbf{Fatal\%} & \textbf{Broken\%} & \textbf{NegDiscr\%} \\
\midrule
gpt-5-mini & 25 & 0.0 & 3.5 & 7.0 \\
gemini-2.5-flash & 24 & 4.2 & 5.0 & 9.8 \\
deepseek-chat & 25 & 0.0 & 7.8 & 9.5 \\
doubao-seed-1-6-flash-250828 & 22 & 0.0 & 12.6 & 8.8 \\
llama-4-maverick & 25 & 8.0 & 10.9 & 7.8 \\
qwen3-next-80b-a3b-instruct & 25 & 8.0 & 14.1 & 12.3 \\
\bottomrule
\end{tabular}

\end{table}

\begin{figure*}[t]
\centering
\includegraphics[width=0.9\linewidth]{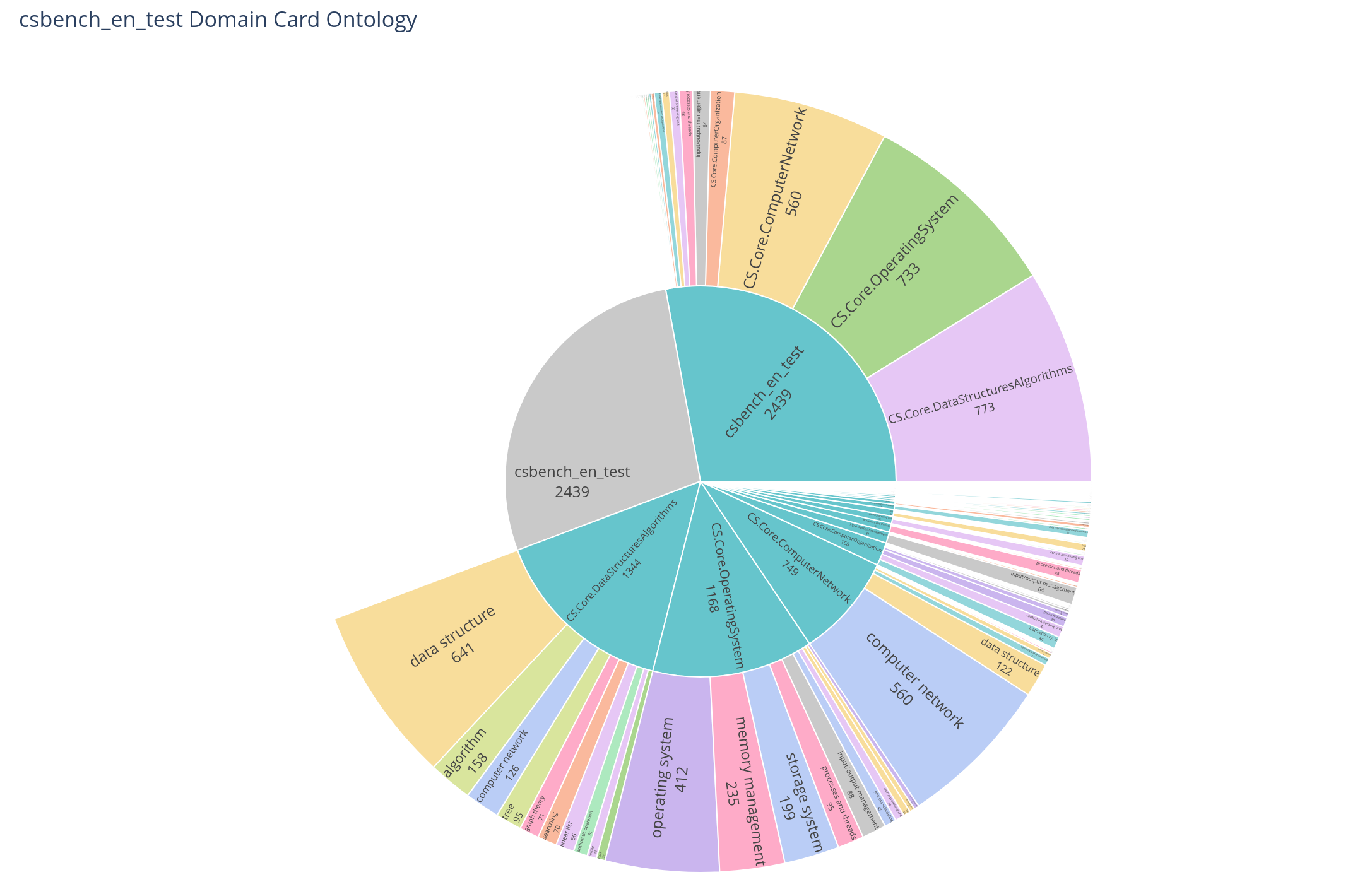}\hfill
\includegraphics[width=0.9\linewidth]{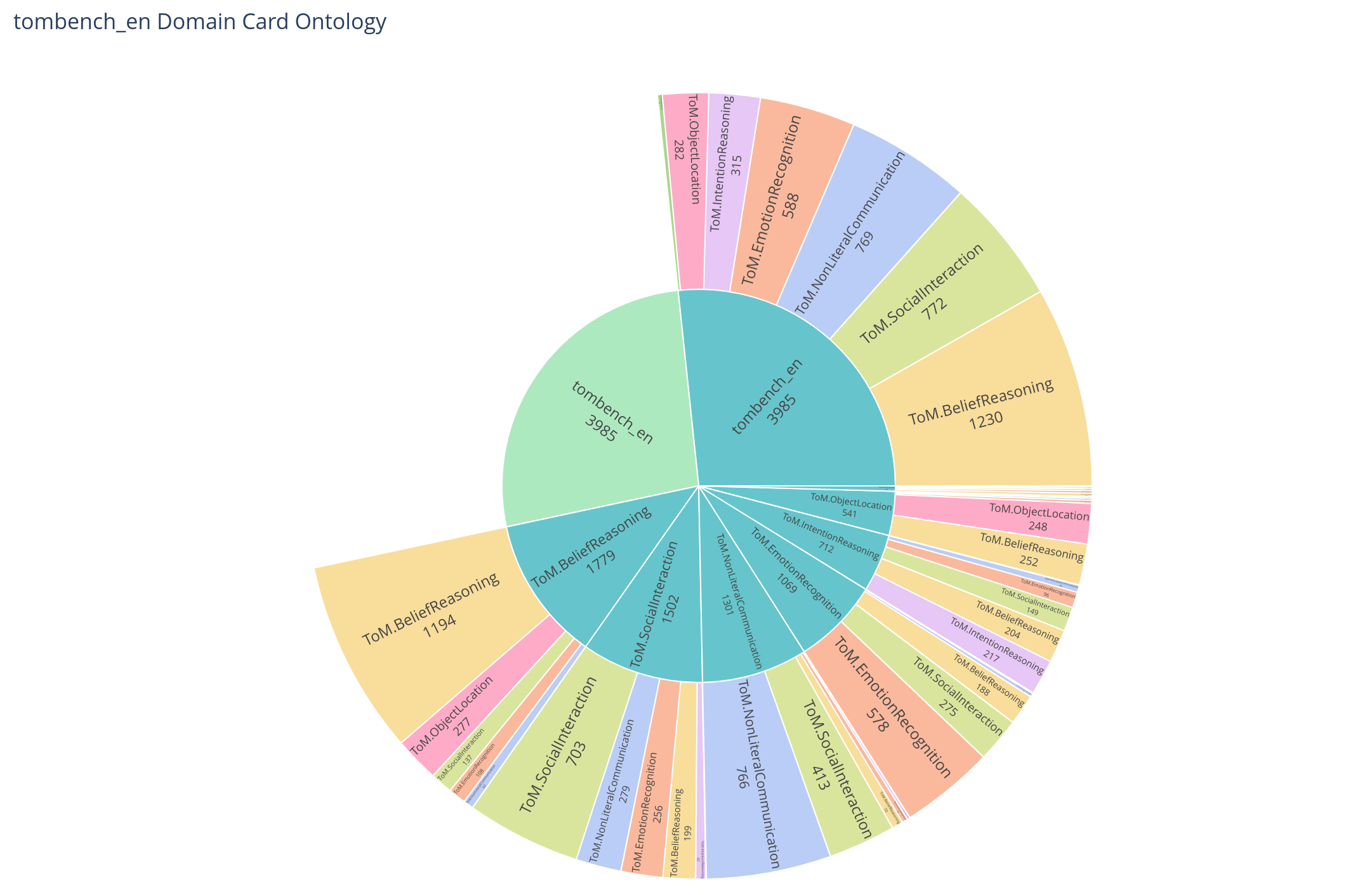}
\caption{\textbf{Example rendered domain cards.} Each plot summarizes the extracted topic ontology (super-parent hierarchy) from a seed benchmark; the domain card also includes representative terms and language/modality constraints.}
\label{fig:domain_card_examples}
\end{figure*}

\subsection{Example Domain Cards}
Figure~\ref{fig:domain_card_examples} shows example rendered domain cards (hierarchy of super-parents) extracted from seed benchmarks. These cards are used as structured conditioning for Stage~2 generation and are released as YAML (plus optional visualizations) in the artifact.

\begin{table*}[t]
\centering
\small
\caption{Designer leaderboard for dataset csbench\_cn.}
\label{tab:designer_leaderboard_csbench_cn}
\begin{tabular}{lrrrrr}
\toprule
Designer & Broken\% & MeanDiscr & NegDiscr\% (all) & Mean $p(\mathrm{correct})$ & \#Core \\
\midrule
qwen3-next-80b-a3b-instruct & 5.0 & 0.434 & 5.7 & 0.803 & 285 \\
gpt-5-mini & 1.7 & 0.391 & 4.3 & 0.865 & 294 \\
doubao-seed-1-6-flash-250828 & 6.7 & 0.366 & 8.0 & 0.833 & 280 \\
gemini-2.5-flash & 3.9 & 0.346 & 8.8 & 0.830 & 274 \\
deepseek-chat & 3.7 & 0.277 & 8.7 & 0.839 & 287 \\
llama-4-maverick & 23.9 & 0.270 & 9.1 & 0.842 & 226 \\
\bottomrule
\end{tabular}

\end{table*}

\begin{table*}[t]
\centering
\small
\caption{Designer leaderboard for dataset medxpertqa\_mm\_stage2\_visualprimed.}
\label{tab:designer_leaderboard_medxpertqa_mm_stage2_visualprimed}
\begin{tabular}{lrrrrr}
\toprule
Designer & Broken\% & MeanDiscr & NegDiscr\% (all) & Mean $p(\mathrm{correct})$ & \#Core \\
\midrule
gpt-5-mini & 2.1 & 0.235 & 7.1 & 0.887 & 275 \\
llama-4-maverick & 15.7 & 0.225 & 8.3 & 0.841 & 253 \\
qwen3-next-80b-a3b-instruct & 13.7 & 0.222 & 6.8 & 0.819 & 253 \\
gemini-2.5-flash & 5.0 & 0.220 & 7.0 & 0.866 & 284 \\
deepseek-chat & 10.3 & 0.128 & 12.7 & 0.859 & 269 \\
doubao-seed-1-6-flash-250828 & 19.6 & 0.105 & 11.3 & 0.766 & 221 \\
\bottomrule
\end{tabular}

\end{table*}

\subsection{File Structure}
We release artifacts in a transparent, analysis-friendly layout:
\begin{itemize}
\item \textbf{Domain cards (YAML)}: one card per variant.
\item \textbf{Stage~2 suites (JSONL)}: per variant and designer, including quota metadata.
\item \textbf{Stage~3 traces (JSONL/JSON)}: per variant, containing responses, scores, and flags.
\item \textbf{Analysis tables/figures (CSV/JSON/SVG)}: aggregated outputs for the paper.
\item \textbf{Code and configs}: scripts and configuration files required to reproduce the pipeline.
\end{itemize}

% \begin{figure*}
% \centering
% \includesvg[width=\textwidth]{wemath.svg}
% \caption{Visual-primed vs. text-only effects in WeMath. Each bar shows $\Delta$ mean discrimination (left) and $\Delta$ mean $p(\mathrm{correct})$ (right) when switching from text-only to visual-primed generation (per designer). Positive $\Delta$ discrimination / negative $\Delta$ $p(\mathrm{correct})$ indicates increased separation and difficulty.}
% \label{fig:visual_delta_wemath}
% \ref{fig:visual_delta_wemath} 
% \end{figure*}
% \section{vistual-primed figure vs text only, visual guide specific analysis}

\section{CSBench-EN Extended Analysis (Claude and Grok Families)}
\label{sec:csbench_en_extended}

CSBench-EN is the only seed suite in our study whose designer set includes two additional model families (\emph{Claude} and \emph{Grok}). To avoid changing the family universe in pooled analyses, we report CSBench-EN-specific observations here.

\paragraph{\textbf{Designer quality with extended families.}}
Table ~\ref{tab:designer_leaderboard_csbench_en} ranks designers by mean discrimination on CSBench-EN. Grok appears among the strongest designers in this suite, producing highly discriminative item sets with moderate invalidity. Claude is competitive but exhibits higher invalidity and a weaker discrimination--validity trade-off than the top designers in this suite.

\begin{table*}[t]
\centering
\small
\caption{Designer leaderboard for dataset csbench\_en.}
\label{tab:designer_leaderboard_csbench_en}
\begin{tabular}{lrrrrr}
\toprule
Designer & Broken\% & MeanDiscr & NegDiscr\% (all) & Mean $p(\mathrm{correct})$ & \#Core \\
\midrule
grok-4-1-fast & 6.8 & 0.307 & 5.1 & 0.880 & 274 \\
llama-4-maverick & 6.4 & 0.298 & 5.4 & 0.902 & 276 \\
gpt-5-mini & 4.3 & 0.291 & 5.1 & 0.896 & 265 \\
qwen3-next-80b-a3b-instruct & 8.1 & 0.242 & 8.1 & 0.846 & 262 \\
claude-sonnet-4-5-20250929 & 9.6 & 0.229 & 7.8 & 0.857 & 244 \\
doubao-seed-1-6-flash-250828 & 10.4 & 0.209 & 6.2 & 0.894 & 232 \\
gemini-2.5-flash & 3.5 & 0.183 & 7.8 & 0.907 & 249 \\
deepseek-chat & 4.8 & 0.173 & 9.2 & 0.882 & 258 \\
\bottomrule
\end{tabular}

\end{table*}

\begin{figure*}[t]
\centering
% Figure file: outputs/paper/svg/fig_family_bias_csbench_en.svg
\includegraphics[width=\linewidth]{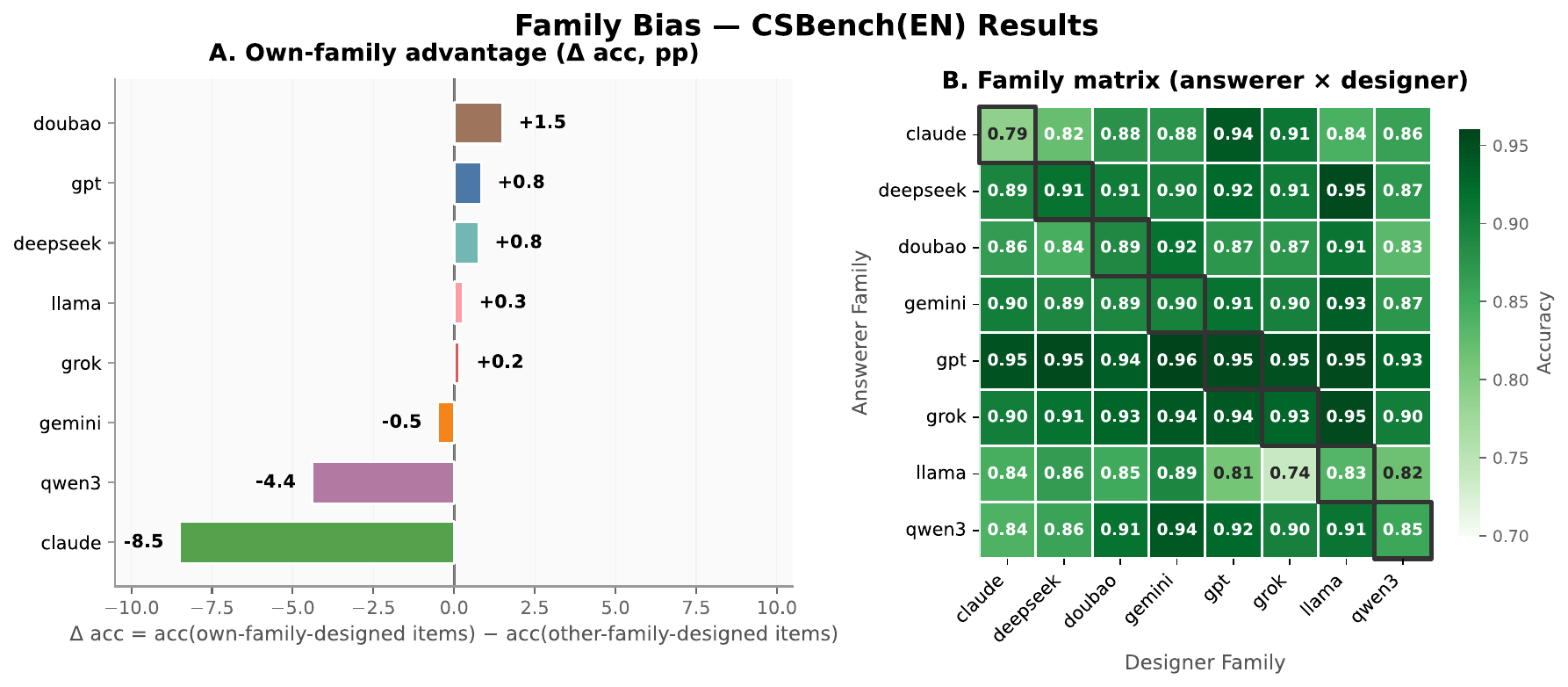}
\caption{Family bias on CSBench-EN with extended designer families. \textbf{Left:} own-family advantage $\Delta$Acc (percentage points), comparing answerer accuracy on items authored by its own family vs.\ other families. \textbf{Right:} family-level designer--answerer accuracy matrix (darker indicates higher accuracy).}
\label{fig:family_bias_csbench_en}
\end{figure*}

\paragraph{\textbf{Family bias is not uniformly ``own-family favorable''.}}
Figure ~\ref{fig:family_bias_csbench_en}  and Table~\ref{tab:family_bias_csbench_en} summarize the own-family vs.\ other-family gap on CSBench-EN. Most families show a small positive gap (higher accuracy on items written by the same family), but this is not universal: Claude and Qwen3 exhibit negative gaps (their answerers perform worse on items authored by their own family than on items from other families). Grok is near-neutral, suggesting limited same-family advantage on this suite. We treat this metric as descriptive: an apparent ``bias'' can also arise when a family authors systematically harder items (e.g., more adversarial or less standard formats), rather than from any preference of the answerer.

\begin{table*}[t]
\centering
\caption{CSBench-EN family bias statistics. Acc$_\mathrm{own}$ / Acc$_\mathrm{other}$ denote answerer-family accuracy on items designed by its own family vs other families; $\Delta$Acc is the difference in percentage points. $n_\mathrm{own}$ and $n_\mathrm{other}$ count evaluated (answerer, item) pairs under each condition.}
\label{tab:family_bias_csbench_en}
\begin{tabular}{lrrrrr}
\toprule
Answerer family & Acc$_\mathrm{own}$ & Acc$_\mathrm{other}$ & $\Delta$Acc (pp) & $n_\mathrm{own}$ & $n_\mathrm{other}$ \\
\midrule
gpt & 0.936 & 0.910 & +2.61 & 529 & 3587 \\
claude & 0.790 & 0.881 & -9.10 & 243 & 1540 \\
gemini & 0.888 & 0.870 & +1.77 & 498 & 3619 \\
grok & 0.927 & 0.926 & +0.15 & 274 & 1786 \\
deepseek & 0.915 & 0.908 & +0.70 & 258 & 1800 \\
qwen3 & 0.864 & 0.895 & -3.14 & 521 & 3594 \\
llama & 0.833 & 0.823 & +1.00 & 552 & 3074 \\
doubao & 0.888 & 0.873 & +1.49 & 232 & 1828 \\
\bottomrule
\end{tabular}

\end{table*}

\paragraph{\textbf{Implication for cross-suite comparison.}}
Because Claude/Grok are absent from the other seed suites, including them in pooled family-bias figures changes the set of families and can complicate comparisons. We therefore recommend (i) reporting the pooled family-bias figure over the intersection of families shared by all suites in the main paper, and (ii) using CSBench-EN as an extended-panel appendix case study with Figure ~\ref{fig:family_bias_csbench_en}  and Table~\ref{tab:family_bias_csbench_en}.

% Appendix section: Prompt templates (Stage 1--3).
% This file is meant to be \input{}'d from the appendix after \appendix.

\section{Prompt Templates}
\label{app:prompt_templates}

This appendix documents the \emph{templates} of prompts used in our three-stage pipeline. We intentionally report prompt templates with placeholders (rather than dataset- or batch-specific instantiated prompts) to keep the paper concise while preserving methodological clarity. The full, instantiated prompts for every dataset/variant and run are released in the accompanying artifact.

\subsection{Stage 1: Domain Card Schema (Knowledge Anchoring)}
\label{app:prompt_templates_stage1}

Stage~1 produces a structured \emph{domain card} (YAML) that anchors Stage~2 generation to a controlled knowledge representation. We include the domain-card schema skeleton below; the extraction procedure and any dataset-specific heuristics are provided in the artifact.

\begingroup\small
\begin{lstlisting}
# Domain card (YAML) schema skeleton
meta:
  dataset: <DATASET_NAME>
  total_items: <INT>
  modality:
    text: <true|false>
    multimodal: <true|false>
    existing_images: <true|false>

ontology:
  - super_parent: <STRING>                # coarse domain (e.g., "OS", "Networks")
    mid_level_parents:
      - label: <STRING>                  # subdomain labels (optional)

glossary:
  - super_parent: <STRING>
    typical_terms: [<TERM_1>, <TERM_2>, ...]

samples:
  - super_parent: <STRING>
    examples:
      - item_id: <STRING>
        question: <STRING>
\end{lstlisting}
\endgroup

\subsection{Stage 2: Controlled Benchmark Generation}
\label{app:prompt_templates_stage2}

Stage~2 uses (i) a \textbf{system prompt} that enforces a strict JSON schema and (ii) a \textbf{per-batch user prompt} that specifies the target domains and quota constraints (difficulty mix, format mix, and standard/adversarial ratio).

\textit{Stage 2 system prompt (template).}

\medskip

\begingroup\small
\begin{lstlisting}
You are an expert exam-question writer for {dataset_display} ({dataset_name}).

Your ONLY job: given a user message describing domains, counts and distributions,
reply with ONE JSON array of question objects and NOTHING else.

Schema (field names and allowed values):

{id, designer_model, source_dataset, super_parent, subdomain,
  design_type in ["standard","adversarial"],
  modality, language,
  question_type in ["mcq_single","mcq_multi","open_ended","structured"],
  question_stem, options, answer, answer_explanation,
  declared_difficulty in ["L1","L2","L3","L4","L5"],
  estimated_time_sec (int),
  uses_visual (bool), visual_instruction}

Every question object MUST include all of these fields.

Conventions:
- designer_model = "{designer_model}"
- source_dataset = "{dataset_name}"
- Defaults if not specified by the user:
  * modality = "text", uses_visual = false, visual_instruction = null
  * language = dataset language (English by default)
- Difficulty: L1..L5 go from very easy recall (L1) to very hard multi-step reasoning (L5, ~90-120s).
- Question types:
  * mcq_single: 4-5 options, exactly 1 correct; answer is a single letter.
  * mcq_multi: 4-5 options, usually 2-3 correct; answer is an array of letters.
  * open_ended / structured: options = []; answer is a short gold text; also provide a concise answer_explanation.
    Keep both answer and explanation strictly under 20 words; avoid multi-paragraph outputs.
- design_type:
  * "standard" = normal fair exam questions.
  * "adversarial" = harder / edge-case but still precise and unambiguous.

Quality rules (for EVERY question):
- Original (no lstlisting copying from examples), domain-correct, solvable from standard knowledge.
- Clear, unambiguous stem and options; plausible distractors, no nonsense.
- Avoid pure trivia and obscure implementation details unless canonical.

Output:
- Return exactly the requested number of questions in a single JSON array.
- No markdown, comments, or prose outside the array.
\end{lstlisting}
\endgroup

\textit{Stage 2 per-batch user prompt (template).}
\begingroup\small
\begin{lstlisting}
Task: generate <TOTAL_QUESTIONS> questions for dataset '<DATASET_DISPLAY_NAME>'.

Target super_parent domains (choose exactly one per question):
- <SUPER_PARENT_1>
  Subdomains: <SUBDOMAIN_1>, <SUBDOMAIN_2>, ...
  Topics: <TERM_1>, <TERM_2>, ...
  Example items: [<ITEM_ID>] <QUESTION_SNIPPET> | ...
- <SUPER_PARENT_2>
  ...

Language: <LANGUAGE_INSTRUCTIONS>
Modality: <MODALITY_INSTRUCTIONS>

Quantitative targets (match as closely as possible):
- standard_questions: <N_STANDARD>
- adversarial_questions: <N_ADVERSARIAL>
- difficulty_targets: {"L1": <...>, "L2": <...>, "L3": <...>, "L4": <...>, "L5": <...>}
- question_type_targets: {"mcq_single": <...>, "mcq_multi": <...>, "open_ended": <...>, "structured": <...>}

ID rules:
- Use IDs of the form "<ID_PREFIX>_qXXX" with zero-padded index.
- Use sequential IDs from <ID_PREFIX>_q001 to <ID_PREFIX>_q<TOTAL_QUESTIONS> with no gaps.

Per-question fixed fields:
- designer_model = "<DESIGNER_MODEL>"
- source_dataset = "<DATASET_NAME>"

Schema reminders (strict):
- answer_explanation must be >= 10 characters.
- For question_type="mcq_multi", answer MUST be a JSON list (e.g., ["A","C"]).
- For question_type="mcq_single", answer MUST be a single string (e.g., "B").
- For question_type="open_ended" or "structured", answer MUST be a string.

Extra guidance (dataset/variant-specific):
- <OPTIONAL_NOTES_1>
- <OPTIONAL_NOTES_2>

Output:
- Return only one single JSON array with exactly <TOTAL_QUESTIONS> question objects and nothing else.
- Output only the JSON array; do not include commentary, summaries, or extra text before/after it.
\end{lstlisting}
\endgroup

\subsection{Stage 3: Answerer Prompts and Quality Judging}
\label{app:prompt_templates_stage3}

Stage~3 performs panel answering with format-specific answer prompts, then applies a two-layer judging procedure: a \textbf{dynamic item-quality judge} to flag ill-posed / ambiguous / gold-incorrect questions, and an \textbf{answer judge} (when needed) to grade open-ended responses against the gold reference.

\textit{Stage 3 answerer prompts (format-specific templates).}
\begingroup\small
\begin{lstlisting}
# mcq_single
SYSTEM: You are taking an exam. Choose exactly one correct option. Do not show any reasoning or steps.
USER:   Question:
        <STEM>

        Options:
        A. <OPT_A>
        B. <OPT_B>
        C. <OPT_C>
        D. <OPT_D>

        Respond with only one capital letter (A, B, C, D) and nothing else.
        No explanation, no justification, no markdown.

# mcq_multi
SYSTEM: You are taking an exam. Some questions may have multiple correct options. Do not show any reasoning or steps.
USER:   Question:
        <STEM>

        Options:
        A. <OPT_A>  ...

        Return the correct options as a JSON array of capital letters.
        Examples: ["A"], ["A","C"], ["B","D","E"].
        Answer with only the JSON array and nothing else.

# structured
SYSTEM: You are taking a structured computer science exam. Follow the requested output format exactly.
USER:   Question:
        <STEM>
        Produce your answer in the requested structure only (e.g., numbered list / table schema / JSON),
        with no additional commentary, reasoning, or prose.

# open_ended
SYSTEM: You are taking a short-answer exam. Give concise, direct answers. Do not include reasoning or steps.
USER:   Question:
        <STEM>
        Answer with ONLY the final answer. No reasoning, no markdown, no extra words.
\end{lstlisting}
\endgroup

\textit{Stage 3 dynamic item-quality judge (Pass-2 template).}
This judge flags problematic items \emph{independent of model performance}, using the stem/options/gold and a small sample of model answers. The judge outputs one of \{\texttt{clean}, \texttt{not\_well\_posed}, \texttt{gold\_incorrect}, \texttt{ambiguous}\} in minified JSON.

\begingroup\small
\begin{lstlisting}
SYSTEM: You are an expert dataset quality judge. Be concise and deterministic. Do not explain your reasoning.
USER:   Question quality check
        Question type: <QUESTION_TYPE>
        Stem: <STEM>
        Options:
        - <OPT_1>
        - <OPT_2>
        ...
        Gold answer: <GOLD_ANSWER>

        Sample model answers:
        - <ANSWERER_1>: <PARSED_ANSWER_1>
        - <ANSWERER_2>: <PARSED_ANSWER_2>

        Return minified JSON only, one line, with key "decision"
        set to one of ["clean","not_well_posed","gold_incorrect","ambiguous"].
\end{lstlisting}
\endgroup

\textit{Stage 3 open-ended answer judge (template).}

When heuristic/exact matching is insufficient (e.g., free-form answers), we call an LLM judge to grade the model answer against the gold reference. The judge outputs JSON only:

\begingroup\small
\begin{lstlisting}
You are an expert grader evaluating a student's answer.

Question:
<STEM>

Gold reference answer:
<GOLD>

Student answer:
<MODEL_ANSWER>

Instructions:
- Be strict on technical accuracy, but tolerant of rephrasing or extra correct detail.
- If uncertain, output "broken_item". Do NOT guess.

Grading rubric:
correct (0.9-1.0): All core ideas present; minor phrasing differences OK.
partially_correct (0.4-0.8): Main idea present, but key details missing/vague/misstated.
incorrect (0.0-0.3): Core concept missing or fundamentally wrong.
broken_item: Question/gold answer is flawed, ambiguous, or factually incorrect.

Output requirements:
- JSON only. No markdown, no extra text.
- "missing" and "errors" must be lists (empty if none).
- Keep items in "missing"/"errors" concise (<=12 words).

{
  "label": "correct" | "partially_correct" | "incorrect" | "broken_item",
  "score": <float>,
  "missing": ["..."],
  "errors": ["..."]
}
\end{lstlisting}
\endgroup

\paragraph{\textbf{Artifact (full prompts).}}
The released artifact contains (i) the exact instantiated prompts for every dataset/variant and run, (ii) all configuration files (model IDs, decoding parameters, batching), and (iii) the domain cards and generated suites used in the experiments.

\newpage

\section{Additional Figures and tables}
\label{app:extra_figs}

\begin{figure*}
  \centering
  \includegraphics[width=\textwidth]{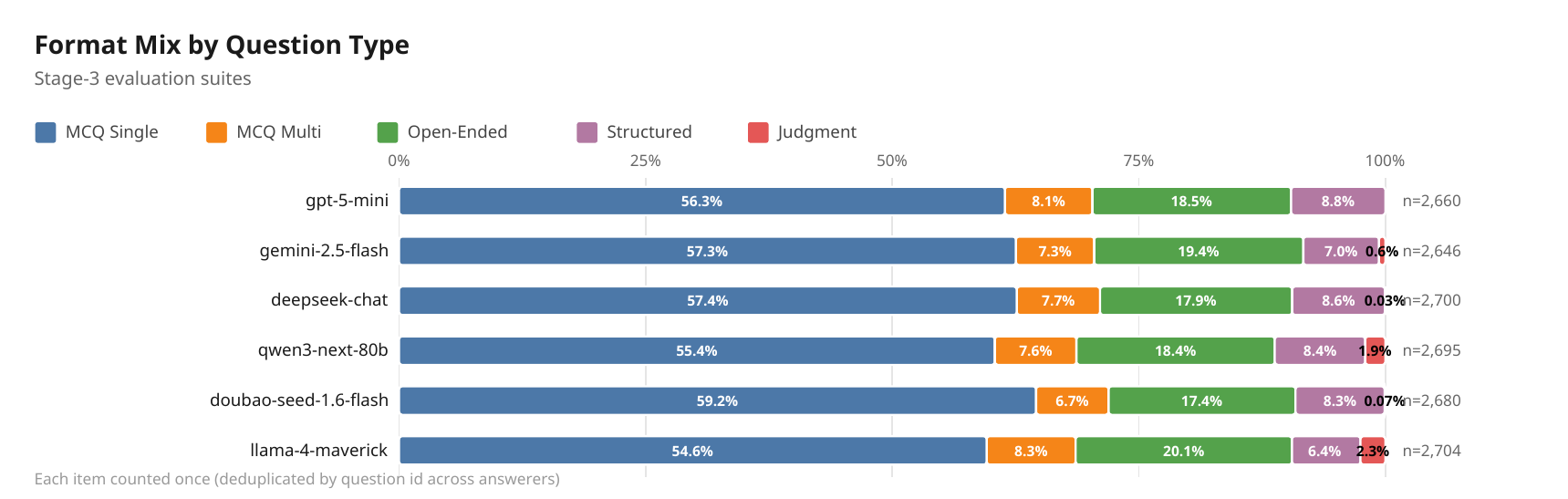}
    \caption{Format distribution across designers (pooled across all generated suites). Each bar shows the proportion of items by question type (mcq\_single, mcq\_multi, open\_ended, structured, judgment). $n$ = total unique items after deduplication.}
  \label{fig:format_mix}
\end{figure*}

% \begin{figure*}[ht]
%     \centering
%     \includesvg[width=\textwidth]{format_mix.svg}
%     \caption{Format distribution across designers (pooled across all generated suites). Each bar shows the proportion of items by question type (mcq\_single, mcq\_multi, open\_ended, structured, judgment). $n$ = total unique items after deduplication.}
%     \label{fig:format_mix}
% \end{figure*}

\begin{figure*}[t]
  \centering
  \includegraphics[width=\linewidth]{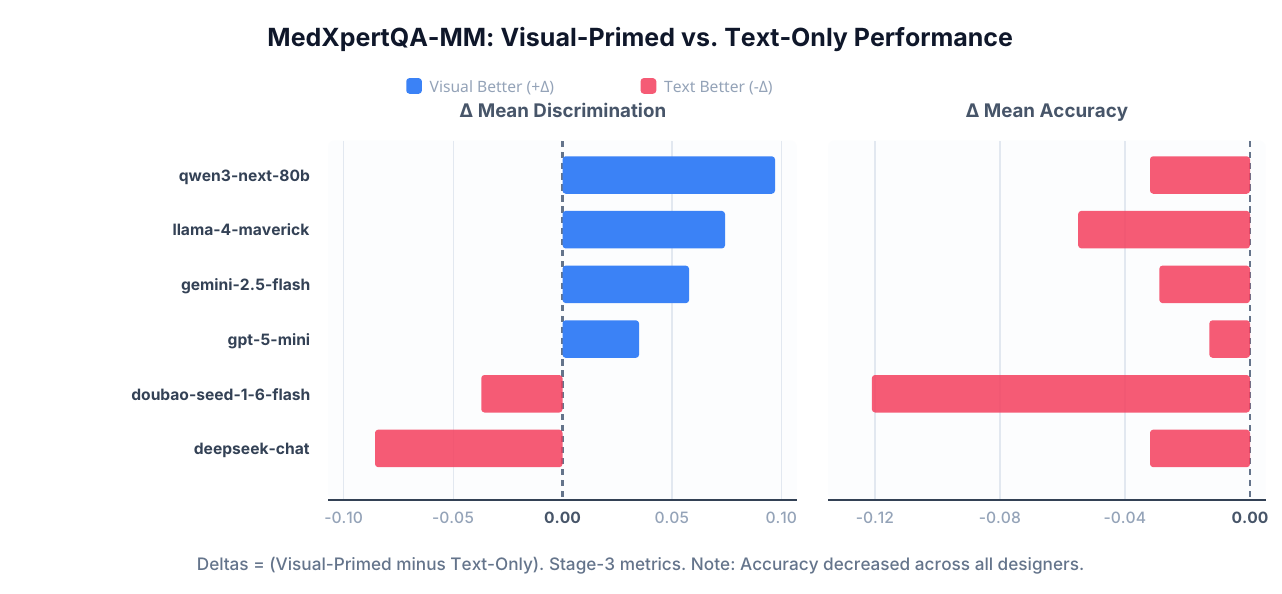}
  \caption{Visual-primed vs.\ text-only effects in MedXpertQA-MM. Each bar shows the change in mean discrimination and mean $p(\mathrm{correct})$ when switching from \texttt{textonly} to \texttt{visualprimed} generation (per designer).}
  \label{fig:visual_delta_med}
\end{figure*}

\begin{figure*}
  \centering
  \includegraphics[width=\textwidth]{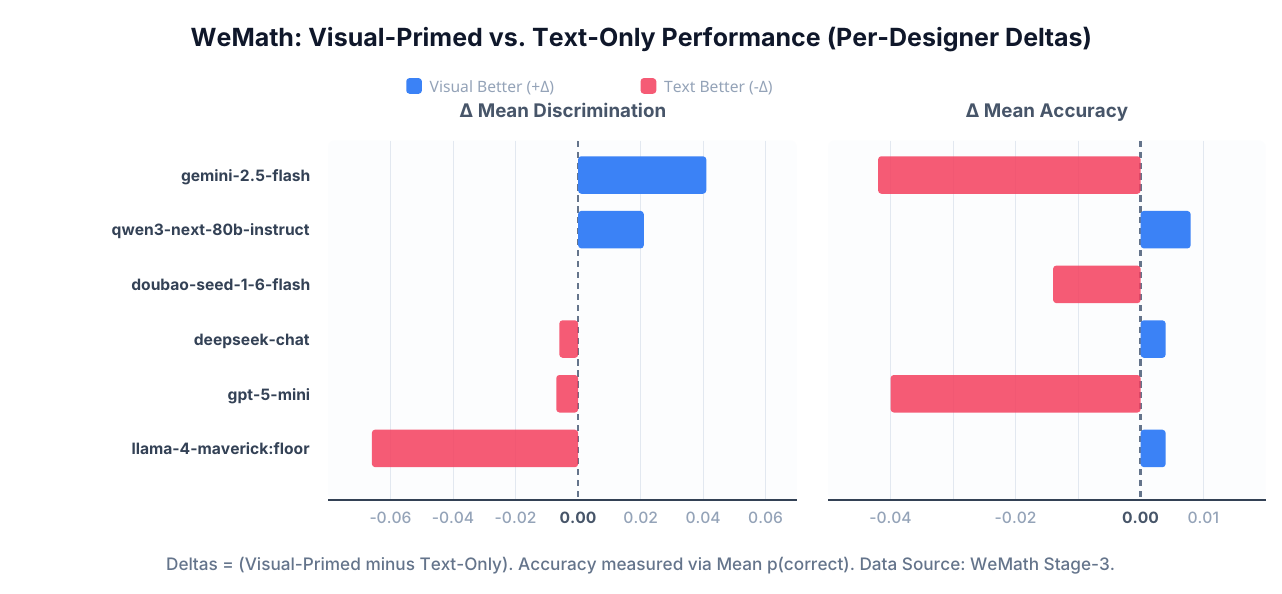}
  \caption{Visual-primed vs.\ text-only effects in WeMath. Each bar shows $\Delta$ mean discrimination (left) and $\Delta$ mean $p(\mathrm{correct})$ (right) when switching from text-only to visual-primed generation (per designer).}
  \label{fig:visual_delta_wemath}
\end{figure*}

% \begin{figure*}

% \centering
% \includesvg[width=\textwidth]{medmm.svg}
% \caption{Visual-primed vs. text-only effects in MedXpertQA-MM. Each bar shows $\Delta$ mean discrimination (left) and $\Delta$ mean $p(\mathrm{correct})$ (right) when switching from text-only to visual-primed generation (per designer). Positive $\Delta$ discrimination / negative $\Delta$ $p(\mathrm{correct})$ indicates increased separation and difficulty.}
% \label{fig:visual_delta_med}

% \end{figure*}
% \begin{figure*}
% \centering
% \includesvg[width=\textwidth]{wemath.svg}
% \caption{Visual-primed vs. text-only effects in WeMath. Each bar shows $\Delta$ mean discrimination (left) and $\Delta$ mean $p(\mathrm{correct})$ (right) when switching from text-only to visual-primed generation (per designer). Positive $\Delta$ discrimination / negative $\Delta$ $p(\mathrm{correct})$ indicates increased separation and difficulty.}
% \label{fig:visual_delta_wemath}
% \end{figure*}

% \begin{figure*}[t]
%   \centering
%   \includesvg[width=\linewidth]{strength_vs_design1.svg}
%   \caption{Answerer strength vs.\ benchmark-design quality for models appearing in both roles.}
%   \label{fig:strength_vs_design}
% \end{figure*}

\begin{figure*}
  \centering
  \includegraphics[width=0.7\textwidth]{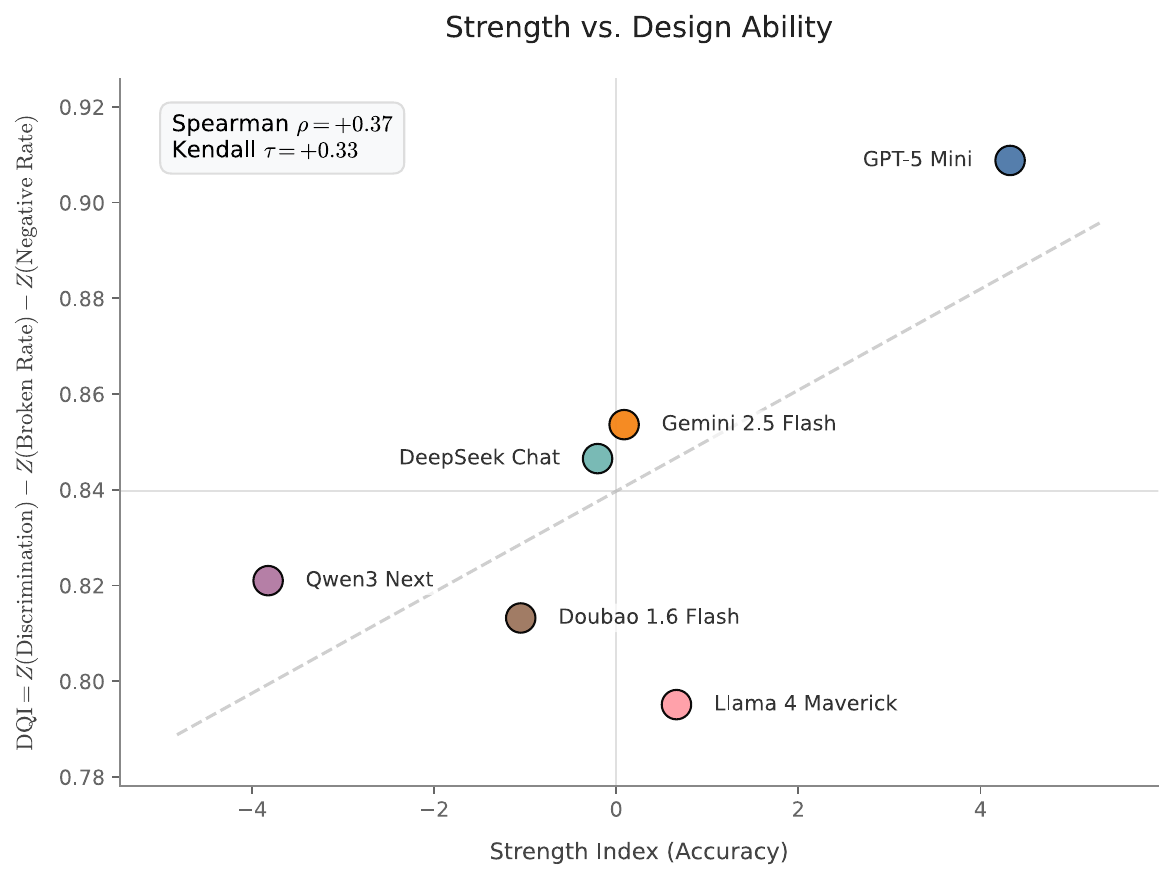}
  \caption{Correspondence between Model Capability and Benchmark Design Quality for models appearing in both roles. The Design Quality Index (DQI) is a composite metric defined as $\text{DQI} = Z(\text{Discrimination}) - Z(\text{Broken Rate}) - Z(\text{Negative Rate})$, which rewards the generation of valid, discriminating questions while penalizing broken or ambiguous items. }
  \label{fig:strength_vs_design}
\end{figure*}

\newpage

\begin{table*}
\centering
\small
\caption{Aggregated designer leaderboard via auto-metrics across all variants. Broken\% represents the fraction of invalid (non-core) items (lower is better for validity); MeanDiscr is the average point-biserial discrimination on core items (higher is better for diagnostic utility); NegDiscr\% (all) denotes the percentage of negatively discriminating items (lower is better to minimize biases); Mean p(correct) proxies item difficulty as the average proportion correct (lower is better for challenge and anti-saturation); \#Core counts retained core items (higher is better for yield). These psychometrics underscore BenchBench's ability to quantify generation tradeoffs, enabling scalable audits of LLM designer prowess.}

\begin{tabular}{lrrrrr}
\toprule
Designer & Broken\% & MeanDiscr & NegDiscr\% (all) & Mean $p(\mathrm{correct})$ & \#Core \\
\midrule
gpt-5-mini &\textbf{3.5} & \textbf{0.301} & \textbf{7.0} & 0.826 & 707 \\
llama-4-maverick & 10.9 & 0.241 & 7.8 & 0.840 & 675 \\
deepseek-chat & 7.8 & 0.213 & 9.5 & 0.812 & 702 \\
doubao-seed-1-6-flash-250828 & 12.6 & 0.209 & 8.8 & \textbf{0.857} & 688 \\
gemini-2.5-flash & 5.0 & 0.205 & 9.8 & 0.836 & \textbf{1281} \\
qwen3-next-80b-a3b-instruct & 14.1 & 0.189 & 12.3 & 0.808 & 1084 \\
\bottomrule
\end{tabular}
\label{tab:designer_leaderboard}
\end{table*}

\end{document}